\documentclass[twoside,11pt]{article}

\usepackage[abbrvbib, preprint]{jmlr2e}
\usepackage[utf8]{inputenc} 
\usepackage[T1]{fontenc}    
\usepackage{lmodern}
\usepackage{booktabs}       
\usepackage{amsfonts}       
\usepackage{amsmath}
\usepackage{amssymb}
\usepackage{mathtools}      
\usepackage{nicefrac}       
\usepackage{microtype}      
\usepackage{xcolor}         
\usepackage{graphicx}
\usepackage[ruled,lined]{algorithm2e}
\usepackage{cancel}
\usepackage{multicol}
\usepackage{cleveref}
\usepackage{url}

\newcommand{\unalign}[1]{%
	\ifmeasuring \else
	\makebox[\ifcase 1\maxcolumn@widths \fi][l]{$\displaystyle#1$}
	\fi
}

\newcommand{\A}{\mathbf{A}}
\newcommand{\polya}{Pólya}
\newcommand{\nystrom}{Nystr\"{o}m}

\renewcommand{\b}{\mathbf{b}}
\newcommand{\B}{\mathbf{B}}

\newcommand{\betadist}{\text{Beta}}
\newcommand{\bigO}{\mathcal{O}}
\newcommand{\bbeta}{\boldsymbol{\beta}}
\newcommand{\bkappa}{\boldsymbol{\kappa}}

\newcommand{\bmu}{\boldsymbol{\mu}}

\newcommand{\bomega}{\boldsymbol{\omega}}
\newcommand{\bOmega}{\boldsymbol{\Omega}}
\newcommand{\bpi}{\boldsymbol{\pi}}
\newcommand{\bPhi}{\varphi_{w}(\X)}
\newcommand{\bPsi}{\boldsymbol{\Psi}}

\newcommand{\bxi}{\boldsymbol{\xi}}

\newcommand{\bSigma}{\boldsymbol{\Sigma}}
\newcommand{\btheta}{\boldsymbol{\theta}}

\newcommand{\E}{\mathbb{E}}
\newcommand{\gammadist}{\mbox{Ga}}
\newcommand{\GP}{\mathcal{GP}}

\newcommand{\N}{\mathcal{N}}
\newcommand{\NB}{\text{NB}}
\newcommand{\poisson}{\text{Poisson}}

\newcommand{\Reals}{\mathbb{R}}
\newcommand{\eye}{\mathbf{I}}
\renewcommand{\d}{\text{d}}
\newcommand{\e}{\mathbf{e}}
\newcommand{\bE}{\mathbf{E}}
\newcommand{\f}{\mathbf{f}}
\newcommand{\F}{\mathbf{F}}
\newcommand{\h}{\mathbf{h}}
\newcommand{\K}{\mathbf{K}}

\newcommand{\p}{\mathbf{p}}
\renewcommand{\P}{\mathbf{P}}
\renewcommand{\r}{\mathbf{r}}
\newcommand{\R}{\mathbf{R}}
\newcommand{\s}{\mathbf{s}}
\renewcommand{\S}{\mathbf{S}}
\newcommand{\V}{\mathbf{V}}
\newcommand{\w}{\mathbf{w}}
\newcommand{\W}{\mathbf{W}}
\newcommand{\x}{\mathbf{x}}
\newcommand{\xp}{\x^{\prime}}
\newcommand{\X}{\mathbf{X}}
\newcommand{\y}{\mathbf{y}}
\newcommand{\Y}{\mathbf{Y}}
\newcommand{\z}{\mathbf{z}}

\newcommand{\zero}{\mathbf{0}}

\newcommand{\Ell}{\mathcal{L}}

\jmlrheading{1}{2023}{1-48}{10/23}{12/23}{zhang23a}{Michael Minyi Zhang, Gregory W. Gundersen, and Barbara E. Engelhardt}

\ShortHeadings{Bayesian Non-linear Latent Variable Modeling via Random Fourier Features}{Zhang, Gundersen, and Engelhardt}
\firstpageno{1}

\begin{document}
	
	\title{Bayesian Non-linear Latent Variable Modeling via Random Fourier Features}
	
	\author{\name Michael Minyi Zhang$^{\star}$ \email mzhang18@hku.hk \\
		\addr
		Department of Statistics and Actuarial Science \\
		University of Hong Kong\\
		Pok Fu Lam, Hong Kong
		\AND
		\name Gregory W. Gundersen$^{\star}$  \email ggundersen@alumni.princeton.edu \\
		\addr Department of Computer Science\\
		Princeton University\\
		Princeton, NJ 08544, USA
		\AND
		\name Barbara E. Engelhardt \email bengelhardt@stanford.edu \\
		\addr Department of Biomedical Data Science\\
		Stanford University\\
		Stanford, CA 94305, USA}
	
\editor{}

\maketitle
\begin{abstract}
The Gaussian process latent variable model (GPLVM) is a popular probabilistic method used for nonlinear dimension reduction, matrix factorization, and state-space modeling. Inference for GPLVMs is computationally tractable only when the data likelihood is Gaussian. Moreover, inference for GPLVMs has typically been restricted to obtaining maximum a posteriori point estimates, which can lead to overfitting, or variational approximations, which mischaracterize the posterior uncertainty. Here, we present a method to perform Markov chain Monte Carlo (MCMC) inference for generalized Bayesian nonlinear latent variable modeling. The crucial insight necessary to generalize GPLVMs to arbitrary observation models is that we approximate the kernel function in the Gaussian process mappings with random Fourier features; this allows us to compute the gradient of the posterior in closed form with respect to the latent variables. We show that we can generalize GPLVMs to non-Gaussian observations, such as Poisson, negative binomial, and multinomial distributions, using our \textit{random feature latent variable model} (RFLVM). Our generalized RFLVMs perform on par with state-of-the-art latent variable models on a wide range of applications, including motion capture, images, and text data for the purpose of estimating the latent structure and imputing the missing data of these complex data sets.
\end{abstract}

\small{$^{\star}$ \textit{both authors contributed equally.}}

\section{Introduction}
A broad category of commonly used machine-learning techniques can be viewed as latent variable models. These methods model hidden structure in data via unobserved or ``latent'' variables. Thus, latent variable models naturally lend themselves to dimension-reduction tasks, since the latent variables can localize the observations in lower-dimensional spaces. Examples of this include matrix factorization techniques, dimension reduction, autoencoders, and state-space models. From a computational perspective, the simplest scenario to fit a probabilistic latent variable model is when the observations are assumed to be Gaussian distributed and when the mapping between the latent and observed variables is assumed to be linear. Many methods fit into this linear-Gaussian framework, such as factor analysis, probabilistic principal component analysis, canonical correlation analysis, and Kalman filters~\citep{roweis1999unifying}. With these linear-Gaussian assumptions, we have closed-form expressions to estimate the statistical parameters and latent variables. However, inference becomes more challenging as we depart from assumptions of normality and linearity, which have many nice mathematical properties with respect to integration. Non-Gaussian and nonlinear models generally do not have closed-form estimates of the latent space.

In this paper, we introduce the \textit{random feature latent variable model} (RFLVM), in which we generalize the Gaussian process latent variable model to a wide variety of non-Gaussian likelihoods by using a random Fourier feature approximation to obtain a computationally tractable inference procedure. Using the random features, we obtain a fairly simple and generalizable Markov chain Monte Carlo (MCMC) sampler for this model which allows us to perform asymptotically exact Bayesian inference in generalized Gaussian process latent variable models. We show that we can easily derive a Gibbs sampler for binomial, multinomial and negative binomial observations using the P\'{o}lya-gamma augmentation scheme and can uncover the latent manifold structure of count data. 

Furthermore, we extend our earlier work in our conference paper by developing a dynamic state space model based off the RFLVM \citep{gundersen2020latent}. We show that, due to the computational tractability from the random features, we can easily extend our RFLVM to model a dynamic latent space, similar to a non-linear state-space model. As a motivating example, consider the problem of modeling neural spike train time series data. In this setting, sets of neurons are recorded as analog signals, which are thresholded into bits and then binned by discretizing time. Thus, these data can be viewed as count data with smooth underlying latent dynamics. A Gaussian assumption is clearly inappropriate, and we must incorporate the time dependent nature of the underlying manifold in the latent variable model as well. Our framework allows for exact posterior exploration of these data using a non-Gaussian, non-linear, state-space model.

We will first begin this paper in~\Cref{sec:background} with an overview of previous work in probabilistic latent variable models, with a specific focus on the Gaussian process-based latent variable models. We then introduce our \textit{random feature latent variable model} in~\Cref{sec:method} and detail the MCMC sampling scheme for exact posterior inference. Next, in~\Cref{sec:experiments} we evaluate our method on a wide range of synthetic and empirical data sets, particularly on neural spike train data sets, to show that we can uncover the latent structure of high dimensional data sets. Lastly, we conclude the paper in~\Cref{sec:conclusion} with a discussion of future directions with our random feature latent variable model.

\section{Background}\label{sec:background}
In this paper, we focus on a simple but broad class of latent variable models taking the following form: our observed data is an $ N \times J $ matrix $ \Y $ with $ N $ observations and $ J $ features; the latent variables are an $ N \times D $ matrix $ \X $ where $ D \ll J$; and these two variables are related through some function of $\X$,

\begin{equation}
\Y = f(\X). \label{eq:basic_lvm}
\end{equation}

\noindent Many latent variable models assume $f(\X)$ is a linear function,

\begin{equation}
\Y = \X \W, \label{eq:linear_lvm}
\end{equation}

\noindent where $f(\X) = \X\W$ for a $ D \times J $ projection matrix $ \W $.

Principal components analysis (PCA), for example, can be viewed through~\Cref{eq:linear_lvm}. In PCA, $\W$ is the solution to an optimization problem of a linear projection of $ \Y $ to a lower-dimensional representation $\X$ which retains the maximum amount of variance from the original data~\citep{hotelling1933analysis,jolliffe2002principal}. Due to linearity, the optimization problem that PCA solves has a simple closed-form solution in the form of the eigenvectors for the sample covariance matrix $\Y\Y^{\top} / N$ corresponding to its $ D $ largest eigenvalues. Non-negative matrix factorization (NNMF) is another latent variable model that fits in the framework of~\Cref{eq:linear_lvm}. In NNMF, the observations, latent variables, and projection matrix are positive-valued \citep{lee1999learning}. Again, due to the linear relationship between the observations and the latent variables, inference in NNMF is fairly simple as we can update the values of $ \W $ and $ \X $ using an alternating least-squares steps.

However, these models are not probabilistic; they have made no statistical assumptions about the latent space or the data-generating process. We can reinterpret~\Cref{eq:basic_lvm} as a probabilistic model by adding a noise term $\bE$ to our basic model,

\begin{equation}
\Y = f(\X) + \bE, \label{eq:basic_proba_lvm}
\end{equation}

\noindent and assuming a distribution on $\X$. For example, \cite{tipping1999probabilistic} formulates PCA as a probabilistic model with Gaussian-distributed latent variables and independent Gaussian-distributed noise, $\e_i \sim \N(0, \sigma_y^2 \eye)$. Since the normal distribution is closed under an affine transformation, we can write probabilistic PCA (PPCA) as

\begin{equation}
\y_i \sim \N(\W\x_i, \sigma^2_{y}\eye), \hspace{1em} \x_i \sim \N(\zero,\eye).
\label{eqn:ppca}
\end{equation}

\noindent In PPCA, the latent space, $\X$, can be marginalized out in closed form:

\begin{equation}
\y_i \sim \N(\zero, \W\W^{\top} + \sigma^2_{y}\eye).
\label{eqn:marginal_ppca}
\end{equation}

\noindent This linear-Gaussian latent variable model has an analytical solution for the maximum likelihood estimate for $\W$, which corresponds to the standard solution in PCA based on eigendecomposition and zero noise \citep{tipping2001sparse,roweis1998algorithms}. PPCA is similar to perhaps the oldest and simplest latent variable model, factor analysis (FA)~\citep{spearman1904general,cattell1945description}, which is equivalent to PPCA but with non-isotropic noise, i.e., $\sigma_y^2 \eye$ is replaced with a diagonal matrix $\bPsi$ in~\Cref{eqn:ppca}.

Additionally, we may place a prior on the mapping weights $ \w_d \sim \N (\zero, \sigma^{2}_{w}) $ and, again, can integrate out the weights and obtain a closed-form representation of the marginal likelihood:

\begin{equation}
\y_i \sim \N(\zero, \sigma^{-2}_{w}\X\X^{\top} + \sigma^2_{y}\eye).
\label{eqn:marginal_ppca2}
\end{equation}

\noindent In fact, \cite{lawrence2005probabilistic} shows that marginalizing the weights and optimizing the latent variables with respect to the marginal likelihood results in a maximum a posteriori (MAP) estimate that is equivalent to the MLE eigenvector solution from PCA, under a particular choice of prior. This formulation of probabilistic PCA under~\Cref{eqn:marginal_ppca2} is crucial to extending the linear model of PPCA to the non-linear, kernel PCA \citep{scholkopf1997kernel}, where the outer product of $ \X\X^{\top} $ is replaced by a kernel matrix $ \K_x$.

\subsection{Gaussian processes}
In the Bayesian setting, a popular choice of prior distribution on the mapping function is the Gaussian process \citep[GP]{williams2006gaussian}, which puts a prior on arbitrary smooth non-linear functions. A GP-distributed function mapping $\x_i$ to $\y_i$,

\begin{equation}
	f \sim \GP(\mu(\cdot), k(\cdot, \cdot)),
\end{equation}

\noindent is defined by its mean function, $\mu(\x_i)$, and covariance function, $k(\x_i,\x_j)$. The defining property of GPs is that a GP-distributed function evaluated at a finite set of points is distributed as a multivariate Gaussian,
\begin{equation}
	f(\x_i) \sim \N(\bmu_x,\K_{xx}).
\end{equation}

\noindent Here, $\bmu_x$ and $\K_{xx}$ denote the mean vector and covariance matrix induced by the mean and covariance functions. This property allows for tractable inference when GPs are used as a prior in non-linear Bayesian generative models, as conditional and marginal distributions of those variables can be written in closed form when the observations have a multivariate Gaussian distribution. 

\subsection{Gaussian process latent variable models}
The Gaussian process latent variable model (GPLVM) provides a Bayesian, probabilistic variant of non-linear latent variable modeling \citep{lawrence2004gaussian,lawrence2005probabilistic} where we let the mean function to be zero, and the observations $\Y$ to be Gaussian distributed:
\begin{equation}
\y_j \sim \N(f_j(\x), \sigma^2_j \eye),
\quad
f_j(\x)\sim \N(\zero, \K_{xx}),
\quad
\x_i \sim \N(0, \sigma^2_x \eye),
\label{eq:gplvm_def}
\end{equation}

\noindent where $\K_{xx}$ is an $N \times N$ covariance matrix defined by a positive definite kernel function $k(\x, \xp)$ and where $f_j(\X) = [f_{j}(\x_1) \dots f_{j}(\x_N)]^{\top}$. Due to conjugacy between the GP prior on $f_{j}$ and Gaussian likelihood on $\y_j$, we can integrate out $f_j$ in closed-form. The resulting marginal likelihood for $\y_j$ :
\begin{equation}
	\y_j \sim \N(\mathbf{0}, \K_{xx} + \sigma^{2}_{j} \eye),
	\label{eq:gplvm_marg}
\end{equation}
From here, we can observe that the marginal likelihood of the GPLVM is identical to the objective function of kernel PCA. Hence, we can interpret kernel PCA as the MLE solution for the GPLVM \citep{lawrence2004gaussian}.

The ability to analytically marginalize the GP-distributed mapping in Equation~\ref{eq:gplvm_marg} allows for computationally tractable posterior inference. Thus, the GPLVM is a popular probabilistic dimension reduction method because it combines flexible nonlinear modeling with computational tractability. We cannot find the optimal $\X$ analytically in the GPLVM, but various approximations have been proposed. We can obtain a MAP estimate by integrating out the GP-distributed maps and then optimizing $\X$ with respect to the posterior using scaled conjugate gradients~\citep{lawrence2004gaussian,lawrence2005probabilistic}, where computation scales as $\bigO(N^3)$. To scale inference, we may use sparse inducing point methods where the computational complexity is $\bigO(NC^2)$, for $C \ll N$ inducing points \citep{lawrence2007learning}. However, these methods only obtain a MAP estimate of the latent space, which is prone to overfitting \citep{damianou2016variational}. 

Instead, we may adopt a variational Bayes approximation of the posterior and minimize the Kullback-Leibler divergence between the posterior and the variational approximation with the latent variables $\X$ marginalized out, in order to infer a posterior distribution, instead of point estimates as in previous work. This variational approach is called a Bayesian GPLVM~\citep{titsias2010bayesian, damianou2016variational}. However, integrating out $\X$ in the approximate marginal likelihood is only tractable when we assume that we have Gaussian observations and when we use an RBF kernel with automatic relevance determination, which limits its flexibility. However, if the observation is no longer assumed to be generated from a Gaussian distribution, we then lose the aforementioned tractability, since we are no longer convolving a Gaussian likelihood with a Gaussian prior. Previous work in non-Gaussian GPLVMs were only capable of obtaining tractability by forming approximations to the posterior distribution, either in the form of a variational or a Laplace approximation. Otherwise, posterior inference in this setting is vulnerable to become trapped in poor local modes. One of the primary reasons why computation is intractable with non-Gaussian likelihoods is because we cannot calculate closed-form gradients of the objective function--the posterior density, with respect to the latent variables.

Further extensions to the basic GPLVM framework have been developed. In the original GPLVM paper, the prior on the latent space, $\X$, is a zero mean Gaussian prior with identity covariance. We can further imbue structure in the latent space prior to better reflect the underlying manifold used to model the data. For example, \cite{lawrence2006local} and \cite{van2009preserving} modify the prior on the latent space so that observations close together in observation space are close together in latent space. From a similar perspective, \cite{urtasun2008topologically} model the latent space with respect to the topological structure of the underlying latent manifold. \cite{urtasun2007discriminative,martens2019decomposing} use label information and covariate data to inform the latent space structure. Beyond additional structure of the latent space, previous work has also incorporated hierarchical components in the generative process to share information across different but associated data sets \citep{lawrence2007hierarchical,kazlauskaite2019gaussian,eleftheriadis2013shared}.

GPLVMs have proven to be a useful statistical model in numerous applications, ranging from single-cell RNA sequencing \citep{verma2020robust,ahmed2018grandprix}, where researchers are interested in studying gene expression patterns in lower dimensions. Previous work have also used GPLVMs successfully in modeling motion capture data and human poses \citep{ek2007gaussian,wang2007gaussian}. Moreover, GPLVMs have been used to model multi-neuron spike train data \citep{wu2017gaussian,she2020neural}, where neuroscientists are interested in modeling a non-linear lower-dimensional representation of the spiking activity. This type of data usually arrive in the form of positive integer-valued counts, where the typical Gaussian assumption of the data becomes inappropriate. This non-Gaussian assumption will result in a loss of computational tractability in the model as the GP-distributed maps are no longer marginalizable in closed form. However, previous work in extending GPLVMs to multinomial and Poisson distributed data used only approximations to the posterior and these approximations are distribution specific \citep{gal2015latent,wu2017gaussian}. However, developing a generalized method for extending GPVLMs to arbitrary data likelihoods still remains a difficult task.

\subsection{Non-linear latent variable models}
Other non-linear dimensionality reduction techniques like stochastic neighbor embedding (SNE) and $t$-SNE represent data with latent variables by minimzing the Kullback-Leibler divergence of the normalized kernel distance of the observed data and the normalized kernel distance of the latent variables \citep{hinton2002stochastic,van2008visualizing}. Uniform manifold approximation and projection (UMAP) preserves the similarity of the observation space and the latent space with respect to the geodesic distances between points, as opposed to Euclidean distances in SNE and $t$-SNE \citep{mcinnes2018umap}. Similarly, locally linear embedding (LLE) uses the same intuition where data that are similar in observation space should be close together in the latent space \citep{saul2003think}. LLE takes a linear combination of an observations neighbors and takes a linear projection of the neighboring local observations into a lower dimension. Another variation of using neighboring data to inform the non-linear dimensional reduction is Isomap, where a distance kernel is calculated for each pair of observations and embeds the data into the latent space by taking the eigenvectors corresponding to the top $ D $ eigenvalues of the pairwise distance matrix \citep{tenenbaum2000global}.

While neural network-based approaches are effective for learning lower dimensional representations, they only provide point estimates without uncertainty quantification. For downstream tasks like prediction, having uncertainty quantification is crucial for correct decision making. In engineering tasks (like tracking a moving target) we need to properly account for a noisy environment so that we may ensure applications of machine learning like autonomous vehicles can correctly identify the position of other vehicles or pedestrians and safely adjust its behavior accordingly. Bayesian models can naturally quantify uncertainty through the posterior distribution. The variational Bayesian autoencoder assumes that the prior distribution for the latent variables is a standard normal and uses a neural network to model the mapping function between the observed space and the latent space. Inference is tractable in this setting as we can optimize with respect to a variational approximation of the marginal likelihood by taking advantage of the \emph{reparameterization trick}, which allows us to construct a gradient-based estimator of the evidence lower bound (ELBO)~\citep{kingma2013auto}. Recent work, like the diffusion probabalistic models assume the latent variables are Gaussian noise propagated through a Markov chain \citep{ho2020denoising,song2021score}, rather than the standard normal assumption on the latent space in the VAE, and has proven to yield state-of-the-art results on a wide range of machine learning benchmark data sets.

\subsection{Latent dynamic variable models}
The basic dynamic latent variable model is the linear state-space model, where the $J$-dimensional observation $\y_i \in \Reals^J$ at time index $i \in [1, \ldots , N]$ is generated by $D \ll J$- dimensional state vector, or latent variable, $\x_i \in \Reals^D$ where the latent dimension $D$ should be much less than the observation dimension, $J$. This vector autoregressive linear system is represented as

\begin{equation}
\x_i = \A \x_{i-1} + \r_i,
\quad
\y_i = \B \x_i + \s_i,
\end{equation}

\noindent where $\A \in \Reals^{D \times D}, \B \in \Reals^{J \times J}$ are state and output transition matrices, and $\r_i \in \Reals^{D} , \s_i \in \Reals^{J}$ are state and output Gaussian noise vectors with mean zero and covariance matrices $\R$ and $\S$, respectively. If the observations and parameters in the linear state-space are assumed to be Gaussian distributed, we can obtain closed-form estimates of the unknown parameters, $(\A,\B,\R,\S)$, using an EM algorithm \citep{ghahramani1996parameter}, or, in the special case of the K\'{a}lman filter, marginalize the parameters and obtain the optimal estimate of the latent space conditioned on the previous observations in closed form \citep{kalman1960new}. However, if the dynamic properties of the process change over time, then we may assume that the state space model follows a switching linear dynamical system \citep[SLDS,][]{oh2008learning}:

\begin{equation}
\x_i = \A_{z_i} \x_{i-1} + \r_i^{(z_i)},
\quad
\y_i = \B \x_i + \s_i,
\quad
z_i \mid z_{i-1} \sim \P_{z_{i-1}},
\end{equation}

\noindent where $z_t = k$, for $k \in \{1, \ldots , K\}$, represents the state indicator variable with latent state switching matrix $\P \in [0,1]^{K \times K}$. Here, the latent transition matrix and the noise parameters $(\A_{z_i}, \r_i^{(z_i)})$ depend on the current hidden state $z_i$ at time $i$. Furthermore, the SLDS model can be extended to the infinite state-space model where the hierarchical Dirichlet process \citep{teh2006hierarchical} is used as the prior over the SLDS parameters, and the number of hidden states is estimated \textit{a posteriori} \citep{fox2011slds}.

\subsection{Non-linear dynamic models}
However, a dynamic model where the latent state space and the observation generating process follow a nonlinear dynamic process may be a more realistic model for many applications because the underlying data generating process is nonlinear. Therefore, functions like the Gaussian process are capable of capturing local dependencies in the latent dynamics between the latent space and the observations through the form of a kernel function. In contrast, linear models assume a constant relationship in the dynamic model.



GPLVMs have been extended to time series data as well, by modifying the prior on the latent variables, $\X$, to incorporate dynamic behavior. The Gaussian process dynamical model (GPDM) assumes a non-linear state-space model~\citep{wang2007gaussian}:

\begin{equation}
\x_i = \A \Phi( \x_{i-1} ) + \r_i, \hspace{1em} \y_i = \B \Phi(\x_i) + \s_i\label{eq:slds}
\end{equation}

\noindent where $\Phi(\cdot)$ represents a basis function. In this model also, if $(\A,\B)$ are assumed to be Gaussian, then we may analytically integrate out these parameters. The resulting marginal likelihood is a GPLVM where the prior on $\X$ is now an autoregressive Gaussian process. Alternatively, we can model the latent dynamic process as a GP regression of the time indices, $i \in \{1, \ldots , N\}$, onto the latent space \citep{lawrence2007hierarchical,damianou2011variational}:

\begin{equation}
p(\X) = \prod_{d=1}^{D} p(\x_d),
\quad
\x_d(i) \sim \N(0, \K_N),
\quad
f_j(\x)\sim \GP(\zero, \K_X),
\quad
\y_j \sim \N(f_j(\x), \sigma^2_j \eye),
\label{eq:dynamic}
\end{equation}

Aside from GP-based models, many other models are common for modeling nonlinear latent dynamic systems, such as the unscented extension of the original Kalman filter \citep{wan2000unscented}, recurrent neural networks (RNNs), and long short-term memory (LSTM) networks, which are extensions of RNNs \citep{hochreiter1997long}. The basic architecture of the recurrent neural network consists of an input $\X_i$, latent hidden layer $\h_i$, model weights $\W$, and an output $\Y_i$:
\begin{equation}
\Y_i = g(\W_{yh} \h_i),
\quad
\h_i = f(\W_{xh}\X_i + \W_{hh} \h_{i-1}),
\end{equation}
\noindent where $f$ is the hidden layer activation function and $g$ is the output layer activation function. Recurrent neural networks can model dynamic structure in the data, but the one-layer RNN is often limited in its expressiveness to approximate different functions. Analogous to how we combine multiple layers of perceptrons to form a deep neural network, we can also form multiple layers of recurrent neural networks to form a deep RNN \citep{pascanu2013construct}. With a deep RNN architecture, we are able to capture highly nonlinear relationships between the observed data and the hidden layers, as well as between the hidden layers themselves. 



\subsection{Random Fourier features}
Kernel-based methods, like the GPLVM, are popular machine learning models because the kernel models nonlinear functions. However, kernel methods scale $ \bigO(N^3) $ in terms of computational complexity and scale $ \bigO(N^2) $ in terms of storage for $ N $ observations. One method to approximate the kernel function is to use random Fourier features (RFFs) by sending the input space through a randomized feature map into $M$ dimensional space, which reduces the computation cost to $ \bigO(NM^2) $ and the storage cost to $ \bigO(NM) $ \citep{rahimi2008random}. Mercer's theorem states that any positive definite kernel function $ k(\cdot,\cdot) $ can be equivalently computed as an inner product of a feature mapping, $ \phi(\cdot) $, between a pair of points so that \citep{mercer1909xvi}:

\begin{equation}
k(\x,\xp) = \left\langle \phi(\x), \phi(\xp) \right\rangle,
\quad
\x,\xp \in \Reals^{D}.
\end{equation}

\noindent If we approximate $ \phi(\cdot) $ with a low-dimensional mapping $ \varphi(\cdot) $ such that $ \varphi(\x_i)^{\top} \varphi(\x_j) \approx \left\langle \phi(\x_i), \phi(\x_j) \right\rangle$, then we can substantially reduce the computational burden of using kernel methods as the memory requirements only scale $O(NM)$ and the computational complexity scales $O(NM^2)$. 

Bochner's theorem states that any continuous shift-invariant kernel function $k(\x, \xp) = k(\x - \xp)$ on $\Reals^D$ is positive definite if and only if $k(\x - \xp)$ is the Fourier transform of a nonnegative measure $p(\w)$. If the kernel is properly scaled, the kernel's Fourier transform $p(\w)$ is guaranteed to be a density \citep{bochner1959lectures}. Let $h(\x) \triangleq \exp(i \w^{\top} \x)$ be a randomized functions which depends on the inputs $\x$ and the random features $\w$, and let $h(\x)^{*}$ denote its complex conjugate:

\begin{equation}
k(\x - \xp)
= \int_{\Reals^D} p(\w) \exp(i \w^{\top} (\x - \xp)) \d\w = \E_{p(\w)}[h(\x)h(\x^{\prime})^{*}].
\label{eq:rffs_mc_int}
\end{equation}

\noindent So $h(\x)h(\x^{\prime})^{*}$ is an unbiased estimate of $k(\x - \xp)$. By dropping the imaginary portion, then $\smash{h(\x) \triangleq \cos(\w^{\top}\x)}$ by Euler's formula. Then by Monte Carlo approximation, we have $k(\x, \xp) \approx \varphi_{w}(\x)^{\top} \varphi_{w}(\x^{\prime})$, where

\begin{equation}
\begin{aligned}
\varphi_{w}(\x) \triangleq \sqrt{\frac{2}{M}} \begin{bmatrix}
\sin(\w_1^{\top} \x),\; \cos(\w_1^{\top} \x)
\\
\vdots
\\
\sin(\w_{M/2}^{\top} \x) ,\; \cos(\w_{M/2}^{\top} \x)
\end{bmatrix}
\label{eq:rff_def},
\quad
\mathbf{w}_m \sim p(\mathbf{w}). 
\end{aligned}
\end{equation}

\noindent We draw $M$ random frequencies from $p(\w)$ to approximate the kernel function. Because the optimal solution to the objective function of a kernel method, $f^{*}(\x)$, is linear in pairwise evaluations of the kernel \citep{kimeldorf1971some}, we can represent $f^{*}(\x)$ as
\begin{equation}
\begin{aligned}
f^{*}(\x)  &= \sum_{n=1}^{N} \alpha_n k(\x_n, \x) = \sum_{n=1}^{N} \alpha_n \langle \phi(\x_n), \phi(\x) \rangle\\
&\approx \sum_{n=1}^{N} \alpha_n \varphi_{w}(\x_n)^{\top} \varphi_{w}(\x) = \bbeta^{\top} \varphi_{w}(\x),
\label{eq:kernel_approx}
\end{aligned}
\end{equation}
\noindent given a reproducing kernel Hilbert space, $\mathcal{H}$. The randomized approximation of this inner product lets us replace expensive calculations involving the kernel with an $M$-dimensional inner product. 

Previous work has only looked at using RFFs to speed up computation \citep{lazaro2010sparse,hensman2017variational}. However, our critical insight regarding RFFs is that by using the random projection of the input space to approximate the kernel function, as seen in~\Cref{eq:kernel_approx}, we can take closed-form evaluations of the likelihood density function. This allows us to use gradient-based optimization techniques to take a MAP estimate or directly take Markov chain Monte Carlo samples. 
Given the representer theorem in~\Cref{eq:kernel_approx}, we can approximate the GPLVM using random features as:
\begin{equation}
\y_j \sim \N_N(\bPhi \bbeta_j, \sigma^{2}_j \eye), \quad
\bbeta_j \sim \N_M(\b_0, \B_0),
\quad
\x_i \sim \N_{D}(\zero, \eye),
\label{eq:gaussian_rflvm_def}
\end{equation}
\noindent which we use as the foundation for our proposed latent variable model.

\subsection{Kernel learning}
In kernel methods, the choice of kernel function, $k(\cdot,\cdot)$, is assumed to already be known \textit{a priori}. However, the choice of kernel function is usually crucial for modeling the behavior of the mapping function, $f(\cdot)$, but is rarely known in most statistical modeling applications. In multiple kernel learning, the kernel function can be estimated through a linear combination of kernels by solving a semidefinite optimization problem \citep{lanckriet2004learning,bach2004multiple}. \cite{yang2015carte} proposed a method for estimating a mixture of kernels using a fast approximation of the kernel using Hadamard matrices which reduce the computational complexity of the kernel machine to $\bigO(N \log J)$. \cite{wilson2013gaussian} proposed a Bayesian variant of kernel learning by introducing a prior over the space of stationary covariance functions in the form of the spectral mixture kernel. Lastly, \cite{oliva2016bayesian} developed a RFF variant of the spectral mixture kernel by placing a Dirichlet process mixture prior on the random frequencies, thereby allowing for a Bayesian non-parametric procedure for kernel learning.

One of the key tools we use to obtain computational tractability in our model is a basis function representation for GPs using a low-rank approximations of a kernel function. The behavior of a GP is typically defined by the choice of kernel used as the covariance function. In the typical implementation of GP-based models, the inverse of the covariance kernel incurs a cubic computational cost with respect to the number of observations. Using random Fourier features, we form a Monte Carlo approximation of the kernel using $M$ random frequencies and can now model a GP-distributed function as a linear function with respect to the random features. In order to take closed-form gradient evaluations of the posterior distribution, we approximate the GP-distributed mappings using random Fourier features (RFFs).  Moreover, this RFF approximation of a GP-distributed function induces a closed form expression for the gradient of the posterior with respect to the latent variables. This is the key insight we need to generalize the GPLVM to the non-Gaussian setting.

\section{Method}\label{sec:method}
We first introduce the Bayesian model with the following data generating process\footnote{In Appendix~\ref{appendix:glossary}, we provide a glossary for the definition of the variables used in this paper.}:
\begin{equation}
\begin{aligned}
\y_j &\sim \Ell \big( g\big( \bPhi \bbeta_j \big), \btheta \big),
&
\btheta &\sim p(\btheta), 
& 
\bbeta_j &\sim \N_{M}(\bbeta_0, \B_0),
\\
\x_n &\sim \N_{D}(\zero, \eye),
&
\w_m &\sim \sum_{k=1}^{\infty} \pi_k \cdot \N_{D}(\bmu_{k}, \bSigma_{k}),
&
\bpi &\sim \mathcal{DP}(\alpha, \mathcal{H}).
\label{eq:model}
\end{aligned}
\end{equation}
Here, $\Ell(\cdot)$ is a likelihood function, $g(\cdot)$ is an invertible link function that maps the real numbers onto the likelihood parameters' support, and $\btheta$ are other likelihood-specific parameters, if they exist (e.g., the dispersion parameter in the negative binomial likelihood). We assume $p(\w)$ is drawn from a Dirichlet process mixture model~\citep{ferguson1973bayesian,antoniak1974mixtures}. For computational tractability, we assign each $\w_m$ in $\smash{\W = [\w_1 \dots \w_{M/2}]^{\top}}$ to a mixture component using variable $z_m$, which is distributed according to a Chinese restaurant process~\citep[CRP,][]{aldous1985exchangeability} with concentration parameter $\alpha$. This prior introduces additional random variables: the mixture means $\smash{\{ \bmu_k \}_{k=1}^{K}}$ and the mixture covariance matrices $\smash{\{ \bSigma_k \}_{k=1}^{K}}$ where $K$ is the number of instantiated clusters in the current sampling iteration. By placing a prior of a Dirichlet process mixtures of Gaussian--inverse Wisharts on the random frequencies, we are able to tractably explore the space of stationary kernels \citep{oliva2016bayesian,wilson2013gaussian}. This choice is in contrast to typical approaches in kernel methods, where the kernel is chosen from a small set of common and tractable kernel functions, such as the radial basis function (RBF), whose Fourier transform corresponds to a Gaussian prior on $ \w_m $. However, the true kernel function in the data generating process is rarely known and should be estimated from the data, hence the flexible choice of the DPMM prior on the kernel.

As seen in~\Cref{eq:kernel_approx}, the optimal approximation of the non-linear function using RFFs is one that is linear with respect to the random features. We use $ \bbeta_j, j \in \{1, \ldots, J\}$ to represent the linear mapping parameters and place a Gaussian prior on $ \bbeta_j $, this allows us to use typical Bayesian linear regression Gibbs sampling updates to take posterior samples of $ \bbeta$.
A typical choice of prior on $\X$ in LVMs assumes \textit{a priori} that there is no dependency structure in the latent space \citep{gundersen2020latent}. In realistic modeling scenarios, we may have more information about the latent space than what a standard Gaussian prior could convey. For example, the discriminative GPLVM incorporates labeled data for the observations to train the model \citep{urtasun2007discriminative}. Dynamic GPLVMs incorporate a temporal dependency by placing a dynamic prior on the latent space, $\X$ \citep{damianou2011variational,lawrence2007hierarchical,wang2007gaussian}. 
Following these dynamic latent GP models, we can place a GP prior on the latent space $\x_d(t) \sim \N(0, \K_T)$, for the dynamic RFLVM, where $\K_T$ is the covariance kernel evaluated on the time indices, $t = 1, \ldots , T$. Previous work in dynamic GPLVMs only achieves tractability by taking advantage of the conjugacy between the GP prior on $\f_j$ and the Gaussian likelihood on $\Y_j$, or they have tailor-made approximate inference strategies for specific likelihoods. However, our random feature-based model flexibly allows for modifications to the prior on the latent variables, $\X$, in a general inference approach.

\subsection{Inference for the random feature latent variable model}
In this section, we present the MCMC sampling steps to perform posteior inference for our proposed latent variable model. We sample the posterior of $\W$ using a Metropolis–Hastings (MH) sampler, where our proposal distribution, $ q(\W) $, is set to the prior. This simplifies the acceptance probability to the ratio of likelihoods:  
\begin{equation}
\begin{aligned}
\w_m^{\star} &\sim q(\W) \triangleq p(\W \mid \z, \bmu, \bSigma), \hspace{1em}
\rho_{\texttt{MH}} = \min \Bigg\{1, \frac{p(\Y \mid \X, \w_m^{\star}, \btheta)}{p(\Y \mid \X, \w_m, \btheta)}\Bigg\}.
\label{eq:w_sampling}
\end{aligned}
\end{equation}
and we sample the latent indicators of the DP mixture, $\z = [z_1 \dots z_{M}]$, using the standard algorithm, ``Algorithm 8'' for sampling for Dirichlet process mixture models \citep{neal2000markov}:
\begin{equation}
\begin{aligned}
p(z_m = k \mid \bmu, \bSigma, \W, \alpha) \sim
\begin{cases}
\frac{n_k^{-m}}{M - 1 + \alpha} \N(\w_m \mid \bmu_k, \bSigma_k) & n_k^{-m} > 0
\\
\frac{\alpha}{M - 1 + \alpha} \int \N(\w_m \mid \bmu, \bSigma) \text{NIW}(\bmu, \bSigma) \text{d}\bmu \text{d}\bSigma & n_k^{-m} = 0.
\end{cases}
\end{aligned}
\label{eq:zm_sampling}
\end{equation}
Given the indicator assignments $\z$ and the random frequencies $\W$, we take advantage of conjugacy between the inverse Wishart-Gaussian prior on the scale-location parameters of the frequencies and Gibbs sample the scale and location parameters \citep{gelman2013bayesian}:
\begin{equation}
\begin{aligned}
\bSigma_k &\sim \mathcal{W}^{-1}(\bPsi_k, \nu_k),
\quad
\bmu_k \sim \N(\mathbf{m}_k, \frac{1}{\lambda_k} \bSigma_k).
\\
\bPsi_k &= \bPsi_0 + \sum_{m:z_m = k}^{M/2} (\w_m - \bar{\w}^{(k)})(\w_m - \bar{\w}^{(k)})^{\top} + \frac{\lambda_0 n_k}{\lambda_0 + n_k} (\w_m - \bmu_0)(\w_m - \bmu_0)^{\top}
\\
\bar{\w}^{(k)} &= \frac{1}{n_k} \sum_{m:z_m=k}^{M} \w_m,
\quad
\nu_k = \nu_0 + n_k, \mathbf{m}_k = \frac{\lambda_0 \bmu_0 + n_k \bar{\w}^{k}}{\lambda_0 + n_k}, \lambda_k = \lambda_0 + n_k.
\end{aligned}
\label{eq:mu_sigma_sampling}
\end{equation}
Finally, we sample the DPMM concentration parameter $\alpha$ by using an augmentation scheme to make sampling $\alpha$ conditionally conjugate with a Gamma prior~\citep{escobar1995bayesian}:
\begin{equation}
\begin{aligned}
\eta &\sim \betadist(\alpha + 1, M), \frac{\pi_{\eta}}{1-\pi_{\eta}} = \frac{a_\alpha + K - 1 }{M(b_\alpha - \log(\eta))},
\;\;K = \left| \{ k: n_k > 0 \} \right|,
\\
\alpha &\sim \pi_{\eta} \gammadist(a_\alpha + K, b_\alpha - \log(h)) + (1 - \pi_{\eta}) \gammadist(a_\alpha + K - 1, b_\alpha - \log(\eta)).
\end{aligned}
\label{eq_alpha_sampling}
\end{equation}
Sampling the posterior of the other likelihood specific parameters (if they exist), $ \btheta $, is the only part of our proposed model that requires a likelihood-specific approach. In some special cases, we have a Gibbs sampling update for the parameter, like an inverse gamma distributed prior on the variance parameter of a Gaussian distribution, but can be sampled generally with gradient-based MCMC algorithms like the Hamiltonian Monte Carlo sampler \citep{duane1987hybrid}.

We cannot obtain an closed form optimal value of $\X$ analytically even in the GPLVM, but various approximations have been proposed. Previous work in GPLVMs infer the latent space by either taking a MAP estimate~\citep{lawrence2005probabilistic} or using a variational approximation~\citep{damianou2016variational}. However, such methods ignore or underestimate the uncertainty in the latent space and could be vulnerable to overfitting. Following prior work in deep Gaussian processes and GPLVMs, we choose to take exact posterior samples of $\X$ using the elliptical slice sampler~\citep[][ESS]{gadd2021pseudo,damianou2013deep,sauer2020active,murray2010elliptical}. In the ESS, we may take posterior draws of any parameter with a Gaussian prior, regardless of the likelihood. The elliptical slice sampler proposes posterior transitions by sampling new states along an ellipse passing through the current parameter state and a draw from the parameter's prior. If the new state is not accepted, then the proposal bracket is shrunk until a new state is accepted. Moreover, ESS does not require any tuning parameters, unlike typical Hamiltonian Monte Carlo samplers, and always transitions to a new state, unlike Metropolis-Hastings samplers. Therefore, we can generalize the sampling process for the latent space despite the choice of likelihood.

In the prior on latent variable $\X$, there are additional parameters governing the dynamic behavior. If we are using the Gaussian process prior on $\X$, then we use an inducing point approximation for the covariance kernel $\K_{xx}$ to reduce the computational complexity of sampling from the prior from $\bigO(N^3)$ to $\bigO(NM^2)$ \citep{snelson2005sparse}. To sample the posteriors of the GP hyperparameters and the locations of the inducing points, we again use the elliptical slice sampler on the Gaussian covariance parameters \citep{murray2010slice}. 

The Gaussian-distributed mapping weights, $\bbeta_j$, in the RFLVM are sampled equivalently to a Bayesian linear model given $\bPhi$. Only the posterior sampling of $ \bbeta_j $ is likelihood dependent, but in cases where we cannot obtain a closed-form sample of the full conditional of $ \bbeta_j $ then we can use the general elliptical slice sampler to take posterior draws (for example, when the likelihood is Poisson-distributed). In other cases, we can again use the elliptical slice sampler to update the mapping weights. If the likelihood is also Gaussian, then we analytically integrate out the weights. For certain count data likelihoods, the mapping weights are equivalent for linear logistic models for which we may use \polya-gamma augmentations to sample $\bbeta_j$ is closed form for binomial \citep{polson2013bayesian}, negative binomial \citep{zhou2012lognormal}, or multinomial likelihoods \citep{chen2013scalable,linderman2015dependent}. For these cases, we may represent the likelihood as equal to 
%
\begin{equation}
\prod_{i=1}^{N}
c(y_{ij})
\frac{(\exp(\varphi_{w}(\x_i) \bbeta_j))^{a(y_{ij})}}{(1 + \exp(\varphi_{w}(\x_i) \bbeta_j))^{b(y_ {ij})}} = 2^{-b_{ij}} e^{\kappa_{ij} \varphi_{w}(\x_i)_{j}} \int_{0}^{\infty} e^{- \omega \varphi_{w}(\x_i)_{j}^2 / 2} p(\omega) \text{d}\omega,
\label{eq:pg_aug_integral}
\end{equation}
where $\kappa_{ij} = a_{ij} - b_{ij}/2$ and $p(\omega) = \text{PG}(\omega \mid b_{ij}, 0)$.  A random variable $\omega$ is \polya-gamma distributed with parameters $b > 0$ and $c \in \Reals$, denoted $\omega \sim \text{PG}(b, c)$, if
\begin{equation}
\omega \stackrel{d}{=} \frac{1}{2\pi^2} \sum_{k=1}^{\infty} \frac{g_k}{(k-1/2)^2 + c^2/(4\pi^2)},
\end{equation}
where $\stackrel{d}{=}$ denotes equality in distribution and $g_k \sim \gammadist(b, 1)$ are independent gamma random variables. \Cref{eq:pg_aug_integral} allows us to rewrite the likelihood as proportional to a Gaussian. We can sample $\omega$ conditioned on $\psi_{ij}$ as $p(\omega \mid \psi_{ij}) \sim \text{PG}(b_{ij}, \psi_{ij})$. This enables convenient, closed-form Gibbs sampling steps of $\bbeta_j$, conditioned on \polya-gamma augmentation variables~$\omega_{ij}$:
\begin{equation}
\begin{aligned}
\omega_{ij} &\mid \bbeta_j \sim \text{PG}( b_{ij}, \varphi_{w}(\x_i)^{\top} \bbeta_j), &  \textbf{V}_{\bomega_j} &= (\bPhi^{\top} \bOmega_j \bPhi + \B_0^{-1})^{-1},  \\
\bbeta_j &\mid \bOmega_j \sim \N(\textbf{m}_{\bomega_j}, \textbf{V}_{\bomega_j}), & \textbf{m}_{\bomega_j} &= \textbf{V}_{\bomega_j} (\bPhi^{\top} \bkappa_j + \B_0^{-1} \bbeta_0),
\end{aligned}
\end{equation}
where $\bOmega_j = \text{diag}([\omega_{1j} \dots \omega_{Nj}])$ and $\bkappa_j = [\kappa_{1j} \dots \kappa_{Nj}]^{\top}$. If we set $a_{ij}=y_{ij}$ and $b_{ij}=1$, then we have a sampler for binomial observations. Alternately, if we set $a_{ij}=y_{ij}$ and $b_{ij}=y_{ij}+r_{j}$, then we have a sampler for negative binomial observations for a dispersion parameter $r_j$. Consider the negative binomial hierarchical model:
\begin{equation}
\begin{aligned}
y_{nj} \sim \NB(r_j, p_{nj}), \qquad r_j \sim \gammadist(a_0, 1/h), \qquad h \sim \gammadist(b_0, 1/g_0).
\end{aligned}
\end{equation}
Then, \cite{zhou2012augment} that showed we can sample $r_j$ as follows:
\begin{equation}
r_j \sim \gammadist\left(L_j, \frac{1}{- \sum_{n=1}^{N} \log(\max(1 - p_{nj}, -\infty))} \right).
\end{equation}
where
\begin{equation}
L_j = \sum_{n=1}^{N} \sum_{t=1}^{\ell_j} u_{n\ell},
\qquad
u_{n\ell} \sim \log(p_{nj}),
\qquad
\ell_j \sim \poisson(-r_j \ln(1-p_{nj})).
\end{equation}

In order to derive a Gibbs sampler for the multinomial likelihood, we first must use the reparameterization of the likelihood \citep{holmes2006bayesian}. We may rewrite the likelihood as
\begin{align}
    \begin{split}
    p(\Y |\X, \bbeta, \W) &= \prod_{i=1}^{N} \frac{\Gamma \left( \sum_{j=1}^{J} y_{ij} +1 \right)}{ \prod_{j=1}^{J} \Gamma \left( y_{ij} +1 \right) } \prod_{j=1}^{J} \left(\frac{\exp \left\{  \varphi_\W(\x_i) \bbeta_j \right\}}{\sum_{j=1}^{J}\exp{\left\{ \varphi_\W(\x_i) \bbeta_j \right\}}}\right) ^{y_{ij}}\\
   &\propto \prod_{i=1}^{N}\prod_{j=1}^{J} \frac{ \left( \exp \left\{ \varphi_\W(\x_i) \bbeta_j - \xi_{ij} \right\} \right)^{y_{ij}}  }{ \left( 1+ \exp \left\{  \varphi_\W(\x_i) \bbeta_j - \xi_{ij} \right\} \right)^{y_{ij}+\sum_{j=1}^{J}y_{ij}} }
    \end{split}
\end{align}
where $\xi_{ij} = \log \sum_{j^{\prime} \neq j} \exp \{ \varphi_\W(\x_i) \bbeta_{j^{\prime}}\} $. By convention and for identifiability purposes, we set $\bbeta_J = 0$. We let $\kappa_{ij} = y_{ij} - \sum_{j=1}^{J}y_{ij}/2$. Now that we have written the likelihood in this form, we may use the \polya-Gamma augmentation trick again where the likelihood is proportional to:
\begin{align}
    \begin{split}
        p(\Y |\X, \bbeta, \W, \bOmega) &\propto \prod_{i=1}^{N}\prod_{j=1}^{J} \exp\Big\{\kappa_{ij}\Big( \varphi_\W(\x_i) \bbeta_j - \xi_{ij}  \Big)  -\frac{\omega_{nj}}{2} \Big( \varphi_\W(\x_i) \bbeta_j - \xi_{ij}  \Big)^2 \Big\}.
    \end{split}
\end{align}
So this gives us a posterior w.r.t. $\bbeta_j$ as
\begin{equation}
p(\bbeta_j \mid \y_j, \X, \bOmega_{j})
\propto
p(\bbeta_j) \prod_{n=1}^{N} \exp\Big\{\bkappa_{i}\Big( \varphi_\W(\x_i) \bbeta_j - \bxi_{j}  \Big)  -\frac{1}{2} \Big( \varphi_\W(\x_i) \bbeta_j - \bxi_{j}  \Big)^{T} \bOmega_j \Big( \varphi_\W(\x_i) \bbeta_j - \bxi_{j}  \Big)  \Big\}.
\end{equation}
We rewrite this into a closed form update as
\begin{equation}
\begin{aligned}
    \bbeta_j \mid \bomega_j &\sim \N(\textbf{m}_{\bomega_j}, \textbf{V}_{\bomega_j})
\end{aligned}
\end{equation}
where
\begin{equation}
\begin{aligned}
    \bOmega_j &= \text{diag}([\bomega_{1j}, \dots, \bomega_{Nj}])
    \\
    \V_{\bomega_j} &= (\bPhi^{\top} \bOmega_j \bPhi + \B_0^{-1})^{-1},
    \\
    \textbf{m}_{\bomega_j} &= \textbf{V}_{\bomega_j} (\bPhi^{\top} (\bkappa_j + \bxi_j^T \bOmega_j ) + \B_0^{-1} \bbeta_0),
    \\
    \bkappa_j &= \y_j - \frac{1}{2} \sum_{j=1}^{J}y_{ij}.
\end{aligned}
\end{equation}
We sample $\bomega_j$ with
\begin{equation}
\begin{aligned}
    \bomega_{j} \mid \bbeta_j &\sim \text{PG}\left( \sum_{j=1}^{J}y_{ij}, \bPhi \bbeta_j - \bxi_j \right).
\end{aligned}
\end{equation}
%
%

\section{Experiments}\label{sec:experiments}\label{sec:experiments:simulated_data}
To evaluate our proposed model, we first examine the ability for the RFLVM to MAP estimate an S-shaped latent space from which we generate a synthetic data set from the prior generating process. Next, we will evaluate the capacity of our latent variable model to place similar observations close together in latent space, given ground truth labels for a variety of empirical data sets. Then, we apply our dynamic RFLVM model on a collection of time series data sets where we will examine how well various static latent variable and dynamic state space models compete in terms of imputing held out data through exact posterior sampling\footnote{Our code is available at \url{https://github.com/michaelzhang01/bayesian-rflvm}.}.


We first evaluate the RFLVM on a simulated data set where the latent space, $\X$ is set to be a 
2-D S-shaped manifold, and the parameters and data simulated from the prior generating process of a GPLVM with an RBF covariance kernel.
For all simulations, we set $N = 500, J = 100$, and $D = 2$. In these experiments, we computed the mean-squared error (MSE) between test set observations, $\Y_{*}$, and predicted observations $\smash{\hat{\Y}_{*}}$, where we held out 20\% of the observations for the test set. 
We evaluated our model's ability to estimate the GP outputs $\smash{f_j(\X) \approx \varphi_{w}(\X) \bbeta_j}$ by comparing the MSE between the estimated $\smash{\varphi_{w}(\hat{\X}) \bbeta_j}$ and the true generating $f_j(\X)$. Lastly, we computed the mean and standard deviation of the MSE 
by running each experiment five times. 

\begin{figure*}[t!]
	\includegraphics[width=\linewidth]{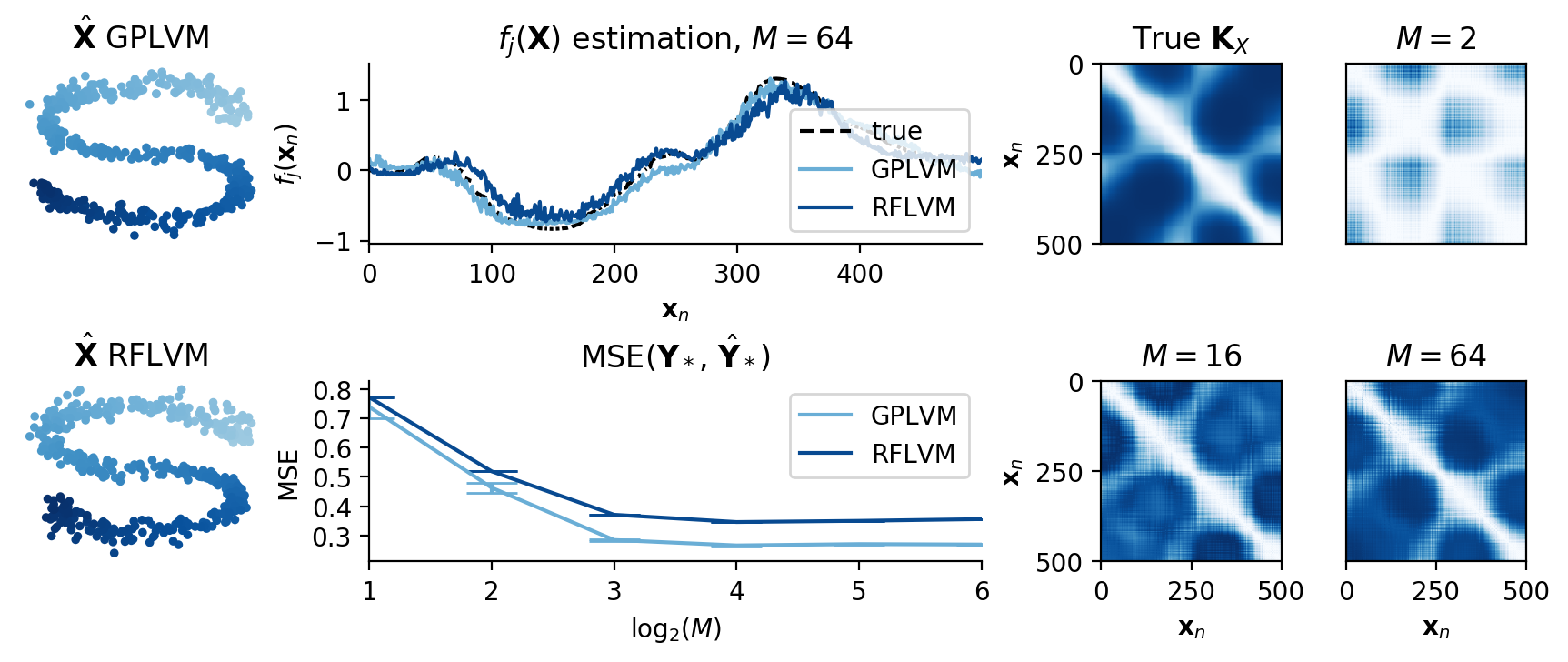}
	\caption{
		\textbf{Simulated data with Gaussian emissions.} (Left) Inferred latent variables for both a GPLVM and Gaussian RFLVM. (Upper middle) Comparison of estimated $f_j(\X)$ for a single feature as estimated by GPLVM and RFLVM. (Lower middle) Comparison of MSE reconstruction error on held out $\Y_{*}$ for increasing $M$, where $M$ is the number of inducing points for GPLVM and random Fourier features for RFLVM. (Right) Ground truth covariance matrix $\K_X$ compared with the RFLVM estimation for increasing $M$.
	}
	\label{fig:gaussian}
\end{figure*}

We compared the performance of a Gaussian RFLVM to the inducing point Bayesian GPLVM, which we will just refer to as the ``GPLVM'' \citep{titsias2010bayesian}\footnote{We fit all of the Bayesian GPLVM experiments using the \texttt{GPy} package~\citep{gpy2014}.}. We ran these experiments across multiple values of $M$, where $M$ denotes the number of random features for the RFLVM and the number of inducing points for the GPLVM. Both models accurately recovered the true latent variable $\X$ and the non-linear maps, $\F$ (Fig.~\ref{fig:gaussian}, upper middle). The GPLVM shows better performance for estimating $\Y_{*}$ than the RFLVM (Fig.~\ref{fig:gaussian}, lower middle). We hypothesize that this could be because \nystrom's method has better generalization error bounds than RFFs when there is a large gap in the eigenspectrum~\citep{yang2012nystrom}, which is the case for $\K_X$. Moreover, we would expect that variational methods like the GPLVM will produce more accurate point estimates for predictions than sampling based methods like RFLVM. However, we see that the RFLVM approximates the true $\K_X$ given enough random features (Fig.~\ref{fig:gaussian}, right), even though it may be less accurate than the GPLVM (Fig.~\ref{fig:gaussian}, lower middle).

Next, we wish to evaluate our model's performance on count-valued data.
We first compared results for simulated count data, sampled from the Poisson RFLVM's prior data generating process, 
against the following benchmarks: PCA, nonnegative matrix factorization~\citep[NMF,][]{lee1999learning}, hierarchical Poisson factorization~\citep[HPF,][]{gopalan2015scalable}, latent Dirichlet allocation~\citep[LDA,][]{blei2003latent}, variational autoencoder~\citep[VAE,][]{kingma2013auto}, deep count autoencoder~\citep[DCA,][]{eraslan2019single}, negative binomial VAE~\citep[NBVAE,][]{zhao2020variational}, and Isomap~\citep{balasubramanian2002isomap}. 
We refer to the Poisson-distributed GPLVM using a double Laplace approximation as \emph{DLA-GPLVM}~\citep{wu2017gaussian}. DLA-GPLVM is designed to model multi-neuron spike train data, and the code\footnote{\url{https://github.com/waq1129/LMT}} initializes the latent space using the output of a Poisson linear dynamical system \citep{macke2011empirical}, and places a GP prior on $\X$. To make the experiments comparable for all GPLVM experiments, we initialize DLA-GPLVM with PCA and assume $\x_n \sim \N_{D}(\zero, \eye)$. 

We refer to our GPLVM with random Fourier features as \emph{RFLVM} and explicitly state the assumed distribution. In Sec.~\ref{sec:experiments:simulated_data}, we use a Gaussian RFLVM with the linear coefficients $\{ \bbeta_j \}_{j=1}^{J}$ marginalized out 
for a fairer comparison with the GPLVM. Since hyperparameter tuning our model on each dataset would be both time-consuming and unfair without also tuning the baselines, we fixed the hyperparameters across experiments. We used 2000 Gibbs sampling iterations with $1000$ burn-in steps, $M = 100$ random features, and a latent dimensionality fixed to $D = 2$. We initialized the number of mixture components to be $K = 20$ and the concentration parameter to be $\alpha = 1$. 
Additionally, we compared results to our own naive implementation of the Poisson GPLVM that performs coordinate ascent on $\X$ and $\F$ by iteratively taking MAP estimates without using RFFs. We refer to this method as \emph{MAP-GPLVM}.

\begin{figure*}[t!]
	\centering
	\includegraphics[width=\linewidth]{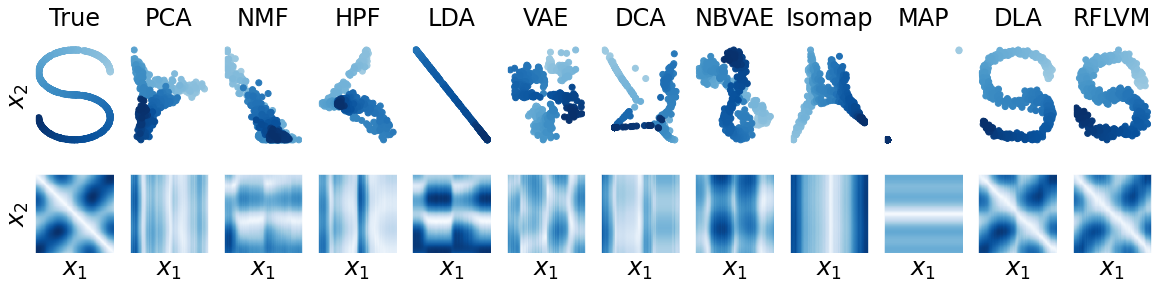}
	\caption{
		\textbf{Simulated data with Poisson emissions.} (Top) True latent variable $\X$ compared with inferred latent variables $\hat{\X}$ from benchmarks (see text for abbreviations) and a Poisson RFLVM. (Bottom) Distance matrices between true $\X$ and $\hat{\X}$ from the above benchmark (darker is farther away).
	}
	\label{fig:poisson}
	\label{fig:illustration}
\end{figure*}

\begin{figure}
	\centering
	\includegraphics[width=1.0\linewidth]{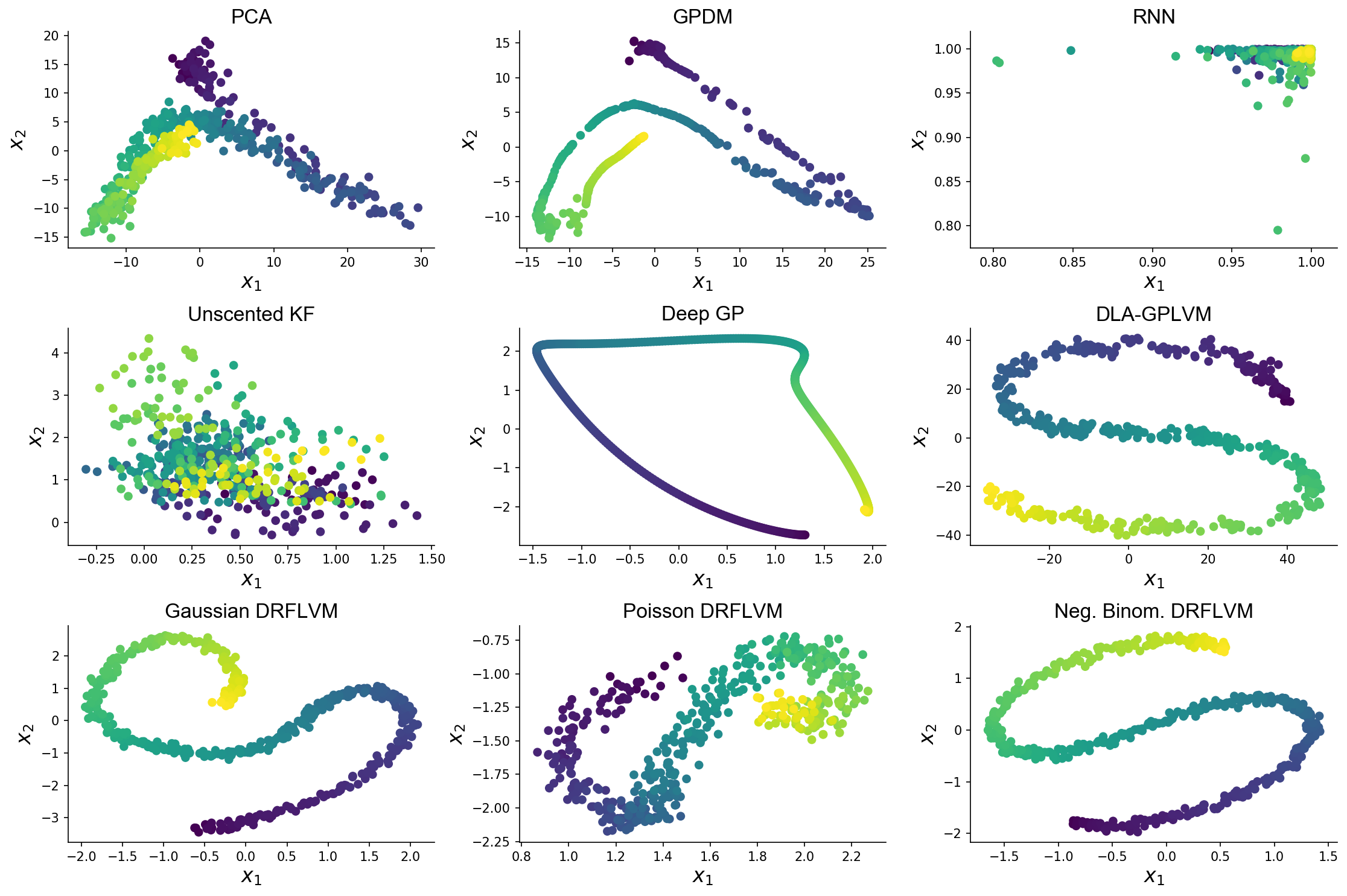}
	\caption{\textbf{Simulated data with Poisson emissions.} Latent dynamic spaces for S-curve across nine methods, labeled in the figure. Color axis refers to time index.}
	\label{fig:scurve}
\end{figure}

We found that the Poisson RFLVM infers a latent variable that is more similar to the true latent structure than other methods (Fig.~\ref{fig:poisson}). Linear methods such as PCA and NMF lack the flexibility to capture this non-linear space, while non-linear but Gaussian methods such as Isomap and VAEs recover smooth latent spaces that lack the original structure. The MAP-GPLVM appears to get stuck in poor local modes~\citep[see][]{wu2017gaussian} because we do not have gradients of the posterior in closed form. Both DLA-GPLVM and RFLVM, however, do have closed-form gradients and approximate the true manifold with similar results. 
Next, we look at a subjective analysis of the inferred latent space for a wide class of state space models in comparison with our dynamic RFLVMs on the synthetic S-curve data sampled from the aforementioend Poisson GPLVM data generating process. In this analysis, we compare our dynamic RFLVMs with PCA, the Gaussian process dynamical model \citep{wang2007gaussian}, a recurrent neural network \citep{hochreiter1997long}, an unscented Kalman filter \citep{wan2000unscented}, and a deep GP \citep{damianou2013deep}\footnote{Implementing a Deep GP as a variant of a dynamic GPLVM came as a suggestion from one of the contributors to \texttt{GPy} \nocite{gpy2014}, due to the fact that the dynamic variational GPLVM is not implemented in the \texttt{GPy} package.}. We see in Fig.~\ref{fig:scurve} that the RFLVMs and the DLA-GPLVM can accurately estimate the S-curve in the latent space, whereas the other dynamic competing methods (GPDM, RNN, UKF, and Deep GP) and PCA cannot properly estimate the S-curve. Again, we can see that correct model specification is important to properly estimating the latent space.

\begin{table*}[!ht]
	\caption{Classification accuracy evaluated by fitting a KNN classifier ($K = 1$) with five-fold cross validation.  Mean accuracy and standard deviation were computed by running each experiment five times.}
	\label{tab:clustering_results}
	\begin{center}
		\resizebox{\textwidth}{!}{%
			\begin{tabular}{lcccccc}
				\toprule
				& PCA & NMF & HPF & LDA & VAE & DCA \\
				\midrule
				Bridges     & $0.8469 \pm 0.0067$ & $\mathbf{0.8664 \pm 0.0164}$ & $0.7860 \pm 0.0328$ & $0.6747 \pm 0.0412$ & $0.8141 \pm 0.0301$ & $0.7093 \pm 0.0317$ \\
				CIFAR-10    & $0.2651 \pm 0.0019$ & $0.2450 \pm 0.0028$ & $0.2516 \pm 0.0074$ & $0.2248 \pm 0.0040$ & $0.2711 \pm 0.0083$ & $0.2538 \pm 0.0178$ \\
				Congress    & $0.5558 \pm 0.0098$ & $0.5263 \pm 0.0108$ & $0.6941 \pm 0.0537$ & $0.7354 \pm 0.1018$ & $0.6563 \pm 0.0314$ & $0.5917 \pm 0.0674$ \\
				MNIST       & $0.3794 \pm 0.0146$ & $0.2764 \pm 0.0197$ & $0.3382 \pm 0.0370$ & $0.2176 \pm 0.0387$ & $\mathbf{0.6512 \pm 0.0228}$ & $0.1620 \pm 0.0976$ \\
				Montreal    & $0.6802 \pm 0.0099$ & $0.6878 \pm 0.0207$ & $0.6144 \pm 0.1662$ & $0.6238 \pm 0.0271$ & $0.6702 \pm 0.0325$ & $0.6601 \pm 0.0997$ \\
				Newsgroups  & $0.3896 \pm 0.0043$ & $0.3892 \pm 0.0042$ & $0.3921 \pm 0.0122$ & $0.3261 \pm 0.0193$ & $0.3926 \pm 0.0113$ & $0.4000 \pm 0.0153$ \\
				Spam        & $0.8454 \pm 0.0037$ & $0.8237 \pm 0.0040$ & $0.8719 \pm 0.0353$ & $0.8699 \pm 0.0236$ & $0.9028 \pm 0.0128$ & $0.8920 \pm 0.0414$ \\
				Yale        & $0.5442 \pm 0.0129$ & $0.4739 \pm 0.0135$ & $0.5200 \pm 0.0071$ & $0.3261 \pm 0.0193$ & $0.6327 \pm 0.0209$ & $0.2861 \pm 0.0659$ \\
				\midrule
				& NBVAE & Isomap & DLA-GPLVM & Poisson RFLVM & Neg. binom. RFLVM & Multinomial RFLVM \\
				\midrule
				Bridges     & $0.7485 \pm 0.0613$ & $0.8375 \pm 0.0240$ & $0.8578 \pm 0.0101$ & $0.8440 \pm 0.0165$ & $\mathbf{0.8664 \pm 0.0191}$ & $0.7984 \pm 0.0102$ \\
				CIFAR-10    & $0.2671 \pm 0.0048$ & $0.2716 \pm 0.0056$ & $0.2641 \pm 0.0063$ & $\mathbf{0.2789 \pm 0.0080}$ & $0.2656 \pm 0.0048$ & $0.2652 \pm 0.0024$ \\
				Congress    & $\mathbf{0.8541 \pm 0.0074}$ & $0.5239 \pm 0.0178$ & $0.7815 \pm 0.0185$ & $0.7673 \pm 0.0109$ & $0.8093 \pm 0.0154$ & $0.6516 \pm 0.0385$ \\
				MNIST       & $0.2918 \pm 0.0174$ & $0.4408 \pm 0.0192$ & $0.3820 \pm 0.0121$ & $0.6494 \pm 0.0210$ & $0.4463 \pm 0.0313$ & $0.3794 \pm 0.0153$ \\
				Montreal    & $0.7246 \pm 0.0131$ & $0.7049 \pm 0.0098$ & $0.2885 \pm 0.0001$ & $\mathbf{0.8158 \pm 0.0210}$ & $0.7530 \pm 0.0478$ & $0.7555 \pm 0.0784$ \\
				Newsgroups  & $0.4079 \pm 0.0080$ & $0.4021 \pm 0.0098$ & $0.3687 \pm 0.0077$ & $\mathbf{0.4144 \pm 0.0029}$ & $0.4045 \pm 0.0044$ & $0.4076 \pm 0.0039$ \\
				Spam        & $\mathbf{0.9570 \pm 0.0045}$ & $0.8272 \pm 0.0047$ & $0.9521 \pm 0.0069$ & $0.9515 \pm 0.0023$ & $0.9443 \pm 0.0035$ & $0.9397 \pm 0.0015$ \\
				Yale        & $0.5261 \pm 0.0346$ & $0.5891 \pm 0.0155$ & $0.4788 \pm 0.0991$ & $\mathbf{0.6894 \pm 0.0295}$ & $0.5394 \pm 0.0117$ & $0.5441 \pm 0.0059$ \\
				\bottomrule
			\end{tabular}
		}
	\end{center}
\end{table*}

\subsection{Text, image, and time series data}\label{sec:experiments:image}
\begin{figure*}[t!]
	\centering
	\includegraphics[width=\linewidth]{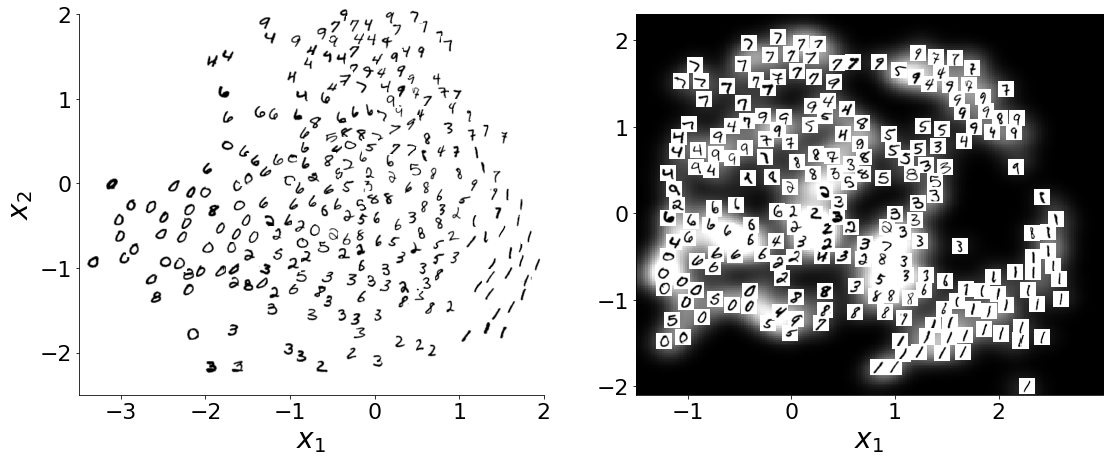}
	\caption{
		\textbf{MNIST digits.}
		Digits visualized in 2-D latent space inferred from DLA-GPLVM (left) and Poisson RFLVM (right). Following \cite{lawrence2004gaussian}, we plotted images in a random order while not plotting any images that result in an overlap. The RFLVM's latent space is visualized as a histogram of 1000 draws after burn-in. The plotted points are the sample posterior mean.
	}
	\label{fig:mnist}
\end{figure*}

\begin{figure*}[h!]
	\centering
	\textbf{}\includegraphics[width=1.0\linewidth]{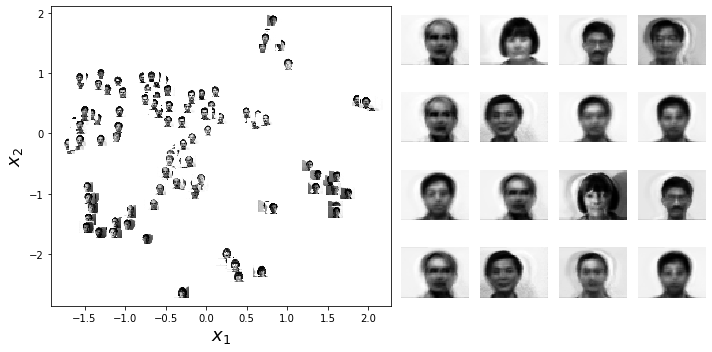}
	\caption{\textbf{Yale face data set.} Face data visualized in 2-D latent space using a Poisson RFLVM (left). Synthetic faces for the Yale dataset sampled from the posterior data generating process using a Poisson RFLVM (right).}
	\label{fig:yale}
\end{figure*}

\begin{figure*}[h!]
	\centering
	\includegraphics[width=1.0\linewidth]{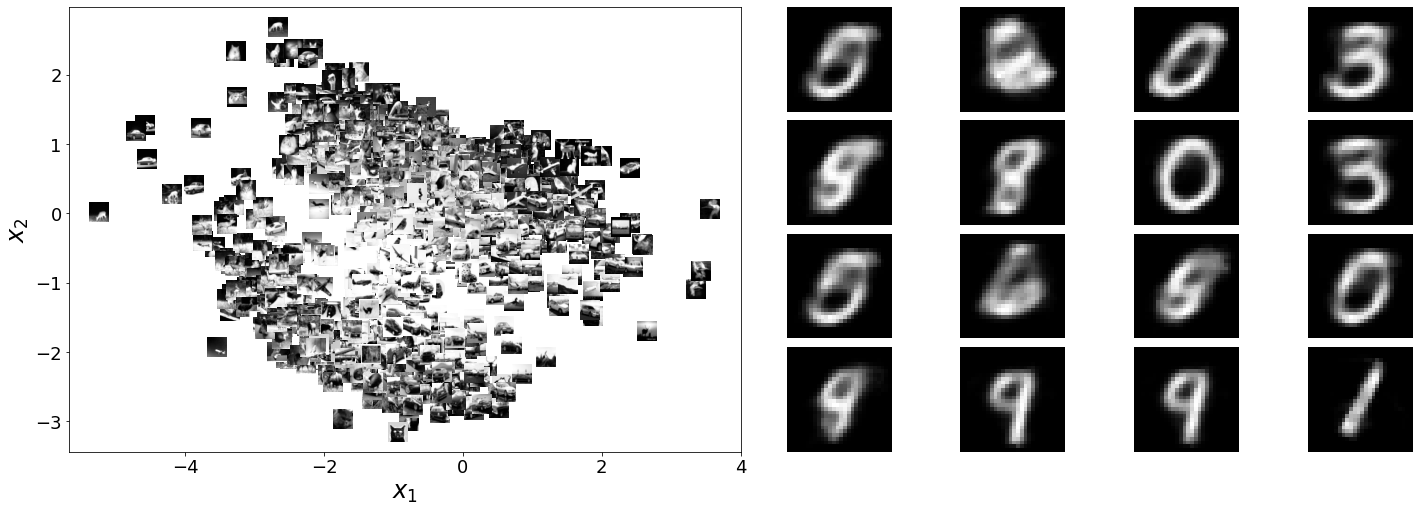}
	\caption{\textbf{CIFAR-10 and MNIST images.} CIFAR-10 image data set visualized in 2-D latent space using a Poisson RFLVM (left). Synthetic digits for MNIST sampled from the posterior data generating process using a Poisson RFLVM (right).}
	\label{fig:cifar10}
\end{figure*}

Next, we examine whether an RFLVM captures the latent space of text, image, and empirical data sets. We hold out the labels and use them to evaluate the estimated latent space using  $K$-nearest neighbors (KNN) classification on $\smash{\hat{\X}}$ with $K=1$. We ran this classification five times using 5-fold cross validation and report the mean and standard deviation of KNN accuracy across five experiments. Across all eight data sets, the Poisson and negative binomial RFLVMs infer a low-dimensional latent variable $\smash{\hat{\X}}$ that generally captures the latent structure as well as or better than linear methods like PCA and NMF~\citep{lee1999learning}. Moreover, adding non-linearity but retaining a Gaussian data likelihood---as with real-valued models like Isomap \citep{tenenbaum2000global}, a variational autoencoder \citep[VAE,][]{kingma2013auto}, and the Gaussian RFLVM, or even using the Poisson-likelihood DLA-GPLVM---perform worse than the Poisson and negative binomial RFLVMs (Tab.~\ref{tab:clustering_results}, Figs.~\ref{fig:mnist}, \ref{fig:yale}, \ref{fig:cifar10}). The point of these results is not that RFLVMs are the best method for every dataset, a spurious claim given ``no free lunch'' theorems~\citep{wolpert1997no}, but rather that our framework allows for the easy implementation of a large number of non-conjugate GPLVMs. Thus, RFLVMs are useful when first performing non-linear dimension reduction on non-Gaussian data. We posit that our improved performance is because the generating process from the latent space to the observations for these data sets is (in part) non-linear, non-RBF, and integer-valued. 

\begin{figure}
	\centering
	\includegraphics[width=.25\linewidth]{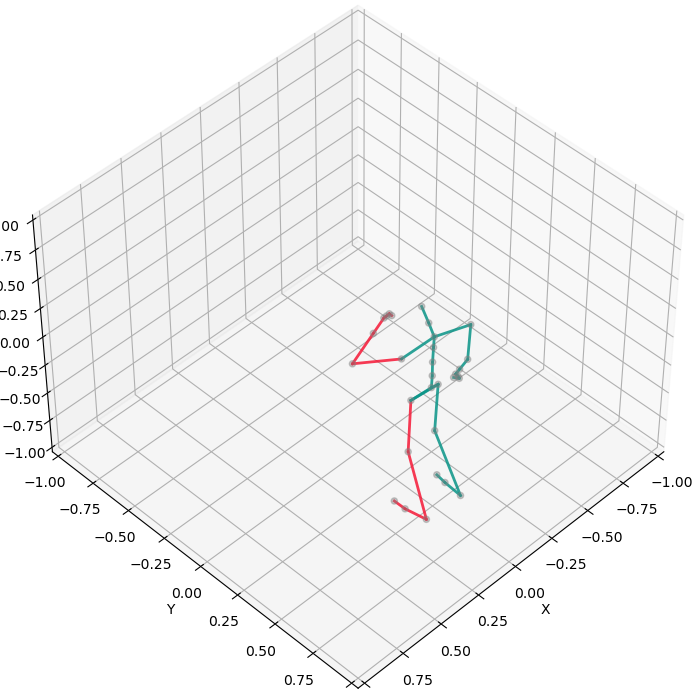}\includegraphics[width=.25\linewidth]{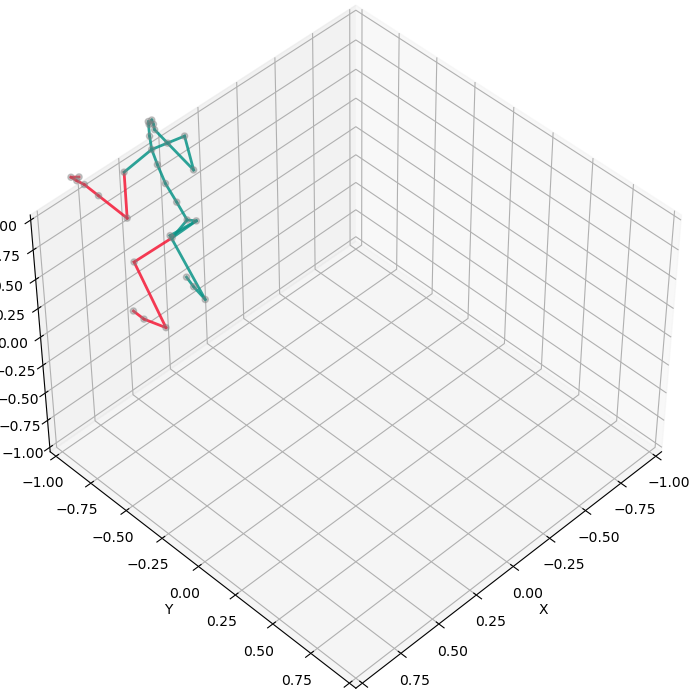}\includegraphics[width=.25\linewidth]{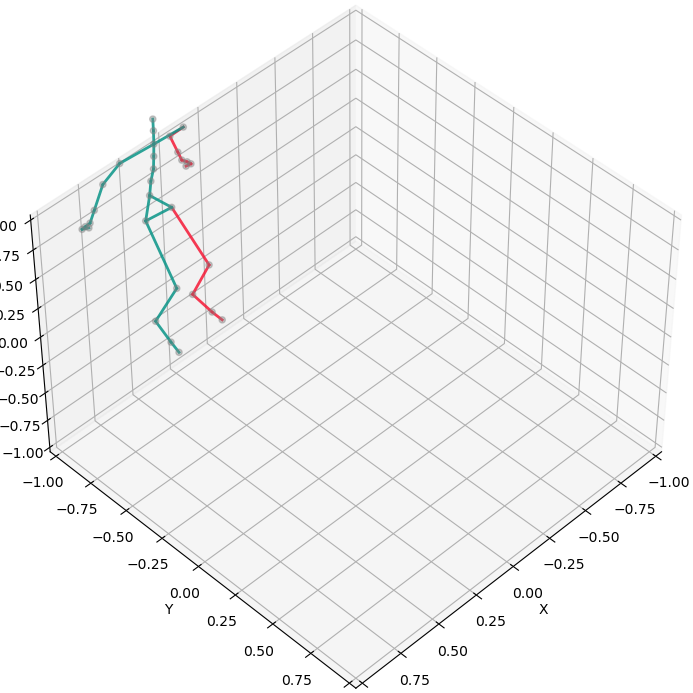}\includegraphics[width=.25\linewidth]{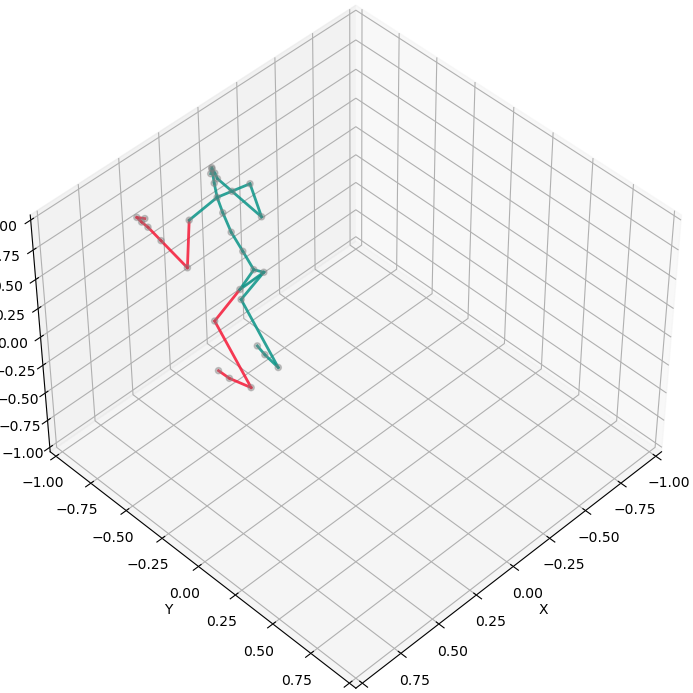}
	\caption{\textbf{CMU motion capture data.} Selected observations from the motion capture data.}
	\label{fig:jumps}
\end{figure}

\begin{figure}
	\centering
	\includegraphics[width=1.0\linewidth]{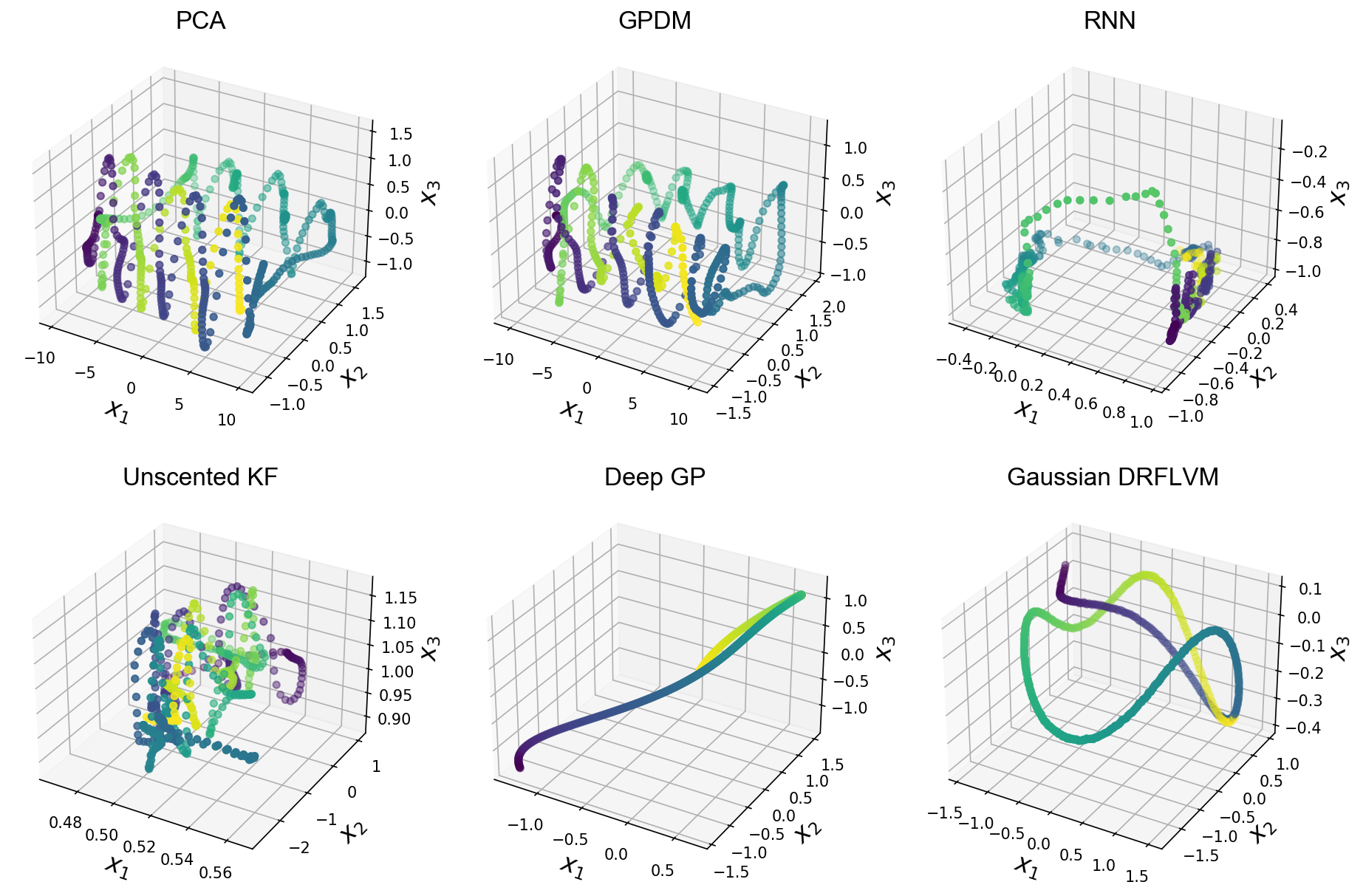}
	\caption{\textbf{CMU motion capture data.} Latent dynamic space for motion capture data across six methods, labeled in the figure. Color axis refers to the time index.}
	\label{fig:cmu}
\end{figure}

As in the Section~\ref{sec:experiments:simulated_data}, we look at a subjective analysis of the latent space estimated from human motion capture data. We compare the latent space of dynamical models on human motion capture data from the CMU Graphics Lab Motion Capture Database. The motion capture data is real-valued data recorded from people performing a variety of actions. In this paper, we look at a recording of someone jumping forward for several leaps, rotating 180 degrees and then jumping several leaps again. In the recorded data that we analyze, the person makes four total laps (see Fig.~\ref{fig:jumps}).  The three dimensional latent space estimated for the motion capture data over a variety of models shows that the GPDM's latent space looks similar to the PCA result, which is what we use to initialize each of the models (Figure~\ref{fig:cmu}). The RNN, UKF, and Deep GP produce latent spaces that are not particularly interpretable, and fail to reflect the dynamics of the observed data. However, results from the dynamic Gaussian RFLVM indeed reflect the four laps of jumps observed in the motion capture data. 

\subsection{Missing data imputation}\label{sec:experiments:missing}
Next, we wish to evaluate the performance of our dynamic RFLVM with some other popular dynamical models in a missing data imputation setting. Here, we randomly hold out 20\% of the $ y_{ij} $ of the observed data, $\Y$, and impute the missing value, $y^{\text{m}}_{ij}$, using its posterior expected value--$ \mathbb{E}\left[y^{\text{m}}_{ij} | \Y^{\text{obs}}, \X, -  \right] $ where $ \Y^{\text{obs}} = \{ y_{ij} : y_{ij} \text{ is observed} \} $. In these experiments, we used the missing data imputation function for the Bayesian GPLVM from \texttt{GPy} to predict the posterior predictive mean of the missing data and implemented a missing value imputer for probabilistic PCA \citep{tipping1999probabilistic}, the GPDM, a linear dynamical model (essentially a GPDM with a linear kernel, $ \K_X = \X\X^T$), a static RFLVM, and the dynamic RFLVM. In Table~\ref{table:ma_results} we see the mean squared error averaged over five trials for the missing data imputation, reported with one standard error. The data sets that we examine in this experiment include the S-curve data set used previously, the synthetic Lorenz attractor typically used in state space model evaluation, the aforementioned CMU data set, and a data set of foreign currency values across time\footnote{\url{https://www.kaggle.com/datasets/brunotly/foreign-exchange-rates-per-dollar-20002019}.}. We use the same experimental settings from the previous sections for the missing data experiment as well. With the exception of the CMU data set, we see that our dynamic RFLVM performs the best on this task in terms of the mean squared error. Thus, we can see that the RFLVM is capable of performing well in a predictive setting, as well as capturing the latent space for a wide variety of synthetic and empirical data sets.

\subsection{Scalability}\label{app:timing}
To assess scalability of RFLVMs, we computed the wall-time in minutes required to fit both RFLVMs and the benchmarks (Table~\ref{tab:timing}). For both the VAE and deep count autoencoder, we trained the neural networks for 2000 iterations (default used in software package\footnote{\url{https://github.com/theislab/dca}}). For DLA-GPLVM, we ran the optimizer for 50 iterations (default used in software package\footnote{\url{https://github.com/waq1129/LMT}}). For RFLVMs, we ran the Gibbs samplers for 100 iterations. While results in Table~\ref{tab:clustering_results} were run for 2000 Gibbs sampling iterations to ensure convergence for all datasets, we found empirically that reducing the number of iterations to 100 did not significantly change the results. All experiments in this section were run on the Princeton University computing cluster, using only CPUs for computation. We find that RFLVMs are indeed slower than most methods, but not substantially so. For example, on the CIFAR-10 dataset, a VAE takes 23.7 minutes, while a Poisson RFLVM takes 22.9 minutes, and a negative binomial RFLVM takes 55.7 minutes. The DLA-GPLVM is the slowest, taking 69.8 minutes.
\begin{table*}[t!]
	\caption{Wall-time in minutes for model fitting. Mean and standard error were computed by running each experiment five times.}
	\label{tab:timing}
	\begin{center}
		\resizebox{\textwidth}{!}{%
			\begin{tabular}{lcccccc}
				\toprule
				& PCA & NMF & HPF & LDA & VAE & DCA \\
				\midrule
				Bridges     & $0.0186 \pm 0.0005$ & $0.0182 \pm 0.0012$ & $0.0273 \pm 0.0002$ & $0.0528 \pm 0.0067$ & $1.8193 \pm 0.0708$ & $0.5740 \pm 0.0255$ \\
				CIFAR-10    & $0.4398 \pm 0.0743$ & $0.4151 \pm 0.0123$ & $1.0894 \pm 0.0500$ & $0.8674 \pm 0.0199$ & $23.6770 \pm 0.3789$ & $1.1341 \pm 0.0540$ \\
				Congress    & $0.0244 \pm 0.0002$ & $0.0245 \pm 0.0007$ & $0.7296 \pm 0.0824$ & $0.0846 \pm 0.0221$ & $4.2919 \pm 0.0539$ & $0.5448 \pm 0.0134$ \\
				MNIST       & $0.2368 \pm 0.0064$ & $0.2522 \pm 0.0273$ & $1.0004 \pm 0.1880$ & $0.3264 \pm 0.0237$ & $15.3385 \pm 1.8402$ & $0.8719 \pm 0.0618$ \\
				Montreal    & $0.0171 \pm 0.0008$ & $0.0164 \pm 0.0001$ & $0.0523 \pm 0.0350$ & $0.0632 \pm 0.0065$ & $2.0585 \pm 0.0947$ & $0.5028 \pm 0.0120$ \\
				Newsgroups  & $0.0219 \pm 0.0006$ & $0.0227 \pm 0.0000$ & $0.1757 \pm 0.0215$ & $0.1163 \pm 0.0344$ & $6.8089 \pm 0.7869$ & $0.8551 \pm 0.0527$ \\
				Spam        & $0.0230 \pm 0.0004$ & $0.0235 \pm 0.0012$ & $0.3039 \pm 0.0419$ & $0.1262 \pm 0.0381$ & $6.8448 \pm 0.7796$ & $0.7146 \pm 0.0453$ \\
				Yale        & $0.0884 \pm 0.0003$ & $0.0984 \pm 0.0064$ & $0.3774 \pm 0.0181$ & $0.1381 \pm 0.0072$ & $5.5177 \pm 0.1645$ & $0.6410 \pm 0.0223$ \\
				\midrule
				& NBVAE & Isomap & DLA-GPLVM & Poisson RFLVM & Neg. binom. RFLVM & Multinomial RFLVM \\
				\midrule
				Bridges     & $0.0867 \pm 0.0157$ & $0.0098 \pm 0.0018$ & $0.5182 \pm 0.0206$ & $0.3318 \pm 0.0135$ & $0.4915 \pm 0.0502$ & $0.5715 \pm 0.0473$ \\
				CIFAR-10    & $2.1002 \pm 0.0594$ & $0.4366 \pm 0.0034$ & $69.7889 \pm 4.2406$ & $22.9299 \pm 1.2624$ & $55.6701 \pm 2.6837$ & $59.8926 \pm 9.9910$ \\
				Congress    & $1.5898 \pm 0.0725$ & $0.0236 \pm 0.0005$ & $45.8584 \pm 22.9771$ & $9.8935 \pm 0.1041$ & $20.4514 \pm 0.3995$ & $94.0656 \pm 2.7319$ \\
				MNIST       & $2.1104 \pm 0.1020$ & $0.2148 \pm 0.0019$ & $26.4795 \pm 1.5429$ & $17.8148 \pm 0.0493$ & $33.8967 \pm 4.1385$ & $74.3100 \pm 2.1778$ \\
				Montreal    & $0.0819 \pm 0.0009$ & $0.0080 \pm 0.0001$ & $0.8723 \pm 0.0237$ & $0.5006 \pm 0.0143$ & $0.9291 \pm 0.0434$ & $0.8769 \pm 0.0376$ \\
				Newsgroups  & $0.7432 \pm 0.0248$ & $0.0721 \pm 0.0008$ & $1088.2659 \pm 35.5089$ & $2.6502 \pm 0.4063$ & $3.2600 \pm 0.0892$ & $2.8393 \pm 0.1525$ \\
				Spam        & $1.8411 \pm 0.0283$  & $0.0795 \pm 0.0036$ & $440.5963 \pm 26.7444$ & $10.6939 \pm 0.4018$ & $17.9958 \pm 2.8573$ & $19.0018 \pm 2.4612$ \\
				Yale        & $0.7931 \pm 0.0589$ & $0.0402 \pm 0.0026$ & $6.7210 \pm 0.1193$ & $9.8992 \pm 0.5530$ & $21.6030 \pm 0.8839$ & $45.4209 \pm 4.4139$ \\
				\bottomrule
			\end{tabular}
		}
	\end{center}
\end{table*}
%
%
%

%
%
%
%
%
%
\begin{table}
	\caption{Mean squared error of imputed held out missing data for time series data. Mean and standard error were computed by running each experiment five times.}
	\centering
	\begin{tabular}{l|cccc}
		& CMU & S-Curve & Lorenz & Forex\\
		\toprule
		GPLVM & $\mathbf{0.071 \pm 0.014}$ & $0.479 \pm 0.094$ & $1.641 \pm 0.205$ & $1.0019 \pm 0.0038$ \\		
		PPCA & $0.084 \pm 0.001$ & $0.761 \pm 0.013$ & $1.227 \pm 0.012$ & $0.3237 \pm 0.0019$ \\
		GPDM & $0.183 \pm 0.008$ & $0.390 \pm 0.009$ & $1.316 \pm 0.016$ & $0.2327 \pm 0.0035$ \\
		LDM & $0.088 \pm  0.001$ & $0.768 \pm 0.013$ & $1.256 \pm 0.005$ & $0.3605 \pm 0.0041$\\
		RFLVM & $0.093 \pm 0.001$ & $0.367 \pm 0.006$ & $1.221 \pm 0.010$ & $0.3289 \pm 0.0023$\\
		DRFLVM & $0.075 \pm 0.001$ & $\mathbf{0.344 \pm 0.007}$ & $\mathbf{1.198 \pm 0.008} $& $\mathbf{0.1317 \pm 0.0021}$
	\end{tabular}
	\label{table:ma_results}
\end{table}
\section{Conclusion}\label{sec:conclusion}
The GPLVM is popular for non-linear latent variable modeling due to its elegant probabilistic formulation, but inference for a non-Gaussian data likelihood has proven to be challenging due to the fact that inference for latent variables are doubly intractable. In this paper, we introduced a Bayesian method for non-linear latent variable modeling that is designed to accommodate a wide spectrum of data likelihoods. Our approach emulates the GPLVM using a random Fourier feature approximation of the covariance kernel. Using the random Fourier features allows us to sample from the posterior of the latent space for a wide variety of data likelihoods. 
We show that our posterior samples effectively learn the latent dynamics in synthetic and empirical time-series data, as well as accurately predict held out missing data compared to popular latent variable and state space models. Currently, we are extending the RFLVM to incorporate sparsity in the latent space as well as to automatically select the latent dimensionality using an Indian buffet process prior \citep{zhang2022sparse}. In future work, we would like to extend our dynamic random feature latent variable model to accommodate non-stationary behavior in the dynamics, and further investigate the problem of modeling neural spike train data where the observations are sparse time-series counts.

\clearpage
\bibliographystyle{apalike}
\bibliography{references}

\clearpage
\appendix
\section{Glossary of Terms}\label{appendix:glossary}
	\begin{table}[h!]
	\[
	\begin{array}{@{}r l}
	&\textbf{Dimensions}
	\\
	N & \text{Number of samples.}
	\\
	J & \text{Number of features.}
	\\
	D & \text{Number of latent features.}
	\\
	M & \text{Number of random Fourier features.}
	\end{array}
	\]
	\end{table}

	\begin{table}[h!]	
	\[
	\begin{array}{@{}r l}

	&\textbf{Data and model}
	\\
	\y_j & \text{$N$ vector for the $j$-th feature for all observations.}
	\\
	\Y & \text{$N \times J$ matrix of observations.}
	\\
	\x_n & \text{$D$ vector, the $n$-th latent variable.}
	\\
	\X & \text{$N \times D$ matrix of latent variables.}
	\\
	k(\cdot, \cdot) & \text{Positive definite kernel function.}
	\\
	\K_X & \text{$N \times N$ matrix after evaluating $k(\x, \xp)$ for every pair in $\X$.}
	\\
	f_j(\cdot) & \text{GP-distributed function for $j$-th feature of $\Y$}
	\\
	f_j(\X) & \text{$N$ vector after applying $f_j(\cdot)$ to every $\x_n$ in $\X$.}
	\\
	\varphi(\cdot) & \text{Randomized approximation of $k(\cdot, \cdot).$}
	\\
	\varphi(\x_n) & \text{$M$ vector of random Fourier features.}
	\\
	\bPhi & \text{$N \times M$ matrix of random Fourier features for each $\x_n$ in $\X$.}
	\\
	\bbeta_j & \text{$M$ vector of coefficients used to simulate $f_j(\X)$ as $\bPhi \bbeta_j$.}
	\\
	\bbeta_0 & \text{Prior mean of $\bbeta_j$.}
	\\
	\B_0 & \text{Prior covariance of $\bbeta_j$.}
	\end{array}
	\]
	\end{table}

	\begin{table}[h!]	
	\[
	\begin{array}{@{}r l}
	&\textbf{Dirichlet Process Mixture of Gaussian-Inverse Wisharts}
	\\
	\w_m & \text{$D$ vector, the $m$-th random Fourier feature.}
	\\
	\W & \text{$M/2 \times D$ matrix of random frequency components.}
	\\  
	\bmu_k & \text{$D$ vector for the $k$-th mixture component's mean.}
	\\
	\bSigma_k & \text{$D \times D$ matrix for the $k$-th mixture component's covariance.}
	\\
	\lambda_k & \text{Posterior number of observations.}
	\\
	\nu_0 & \text{Degrees of freedom for $\bSigma_k$.}
	\\
	\mu_0 & \text{Mean of $\bmu_k$.}
	\\
	\bPsi_0 & \text{Inverse-Wishart scale.}
	\\
	\lambda_0 & \text{Prior observations.}
	\\
	z_m & \text{Scalar mixture component assignment, taking values in $1, \dots, K$.}
	\\
	\z & \text{$M/2$ vector of mixture component assignments.}
	\\
	\alpha & \text{DP concentration parameter.}
	\\
	a_{\alpha} & \text{Gamma shape hyperprior for DP concentration parameter $\alpha$.}
	\\
	b_{\alpha} & \text{Gamma rate hyperprior for DP concentration parameter $\alpha$.}
	\\
	\eta & \text{Augmenting variable to make $(\alpha \mid \eta)$ conditionally conjugate.}
	\end{array}
	\]
\end{table}
\begin{table}[h!]
	\[
	\begin{array}{@{}r l}
	&\textbf{Likelihoods}
	\\
	\mathcal{L}(\cdot) & \text{Likelihood function.}
	\\
	\btheta & \text{Likelihood-specific parameters.}
	\\
	g(\cdot) & \text{Link function mapping $\X$ onto likelihood's support.}
	\\
	\S_0 & \text{Gaussian prior covariance s.t. $\B_0 = \sigma^2 \S_0$}
	\\
	\p_j & \text{$N$ vector of NB probability parameters for $j$-th feature of $\Y$.}
	\\
	\P & \text{$N \times J$ matrix of NB probability parameters.}
	\\
	r_j & \text{Scalar NB dispersion parameter for $j$-th feature of $\Y$.}
	\\
	\mathbf{r} & \text{$J$ vector of NB dispersion parameters.}
	\\
	\bomega_j & \text{$N$ vector of \polya-gamma augmenting variables. }
	\\
	\bOmega & \text{$N \times J$ matrix of \polya-gamma random variables.}
	\\
	a_r & \text{Gamma shape hyperprior for NB dispersion parameter $r_j$.}
	\\
	b_r & \text{Gamma rate hyperprior for NB dispersion parameter $r_j$.}
	\end{array}
	\]
\end{table}

\end{document}